\documentclass[journal]{IEEEtran}

\usepackage{amssymb}
\usepackage{amsfonts}
\usepackage{amsmath}
\usepackage{amsthm}
\usepackage{graphicx}
\usepackage{graphics}
\usepackage{cite}
\usepackage{color}
\usepackage{multirow}
\usepackage{boxedminipage}
\usepackage{booktabs}
\usepackage{algorithm, algpseudocode}
\usepackage{enumitem}
\usepackage{hyperref}

\input{mysymbol.sty}

\begin{document}

%\title{Solving$\hspace{.2cm}$Finite-Horizon$\hspace{.2cm}$MDPs$\hspace{.2cm}$via$\hspace{.2cm}$Low-Rank$\hspace{.2cm}$Tensors}

\title{{Addressing Finite-Horizon MDPs via Low-Rank Tensor Value Approximation}}

\author{Sergio~Rozada,~\IEEEmembership{Member,~IEEE,}
José Luis Orejuela, and
Antonio~G.~Marques,~\IEEEmembership{Senior Member,~IEEE}

\thanks{S. Rozada, J. L. Orejuela  and A. G. Marques are with the Department of Signal Theory and Comms., King Juan Carlos University, Madrid, Spain, sergio.rozada@urjc.es, jl.orejuela.2018@alumnos.urjc.es, antonio.garcia.marques@urjc.es. This work has been partially supported by the Spanish AEI 
(\href{https://doi.org/10.13039/501100011033}{10.13039/501100011033}) under Grants PID2022-136887NB-I00, TED2021-
130347B-I00, and the Community of Madrid via the Ellis Madrid Unit and grant TEC-2024/COM-89. Minor edits of this document were made with the assistance of ChatGPT.}}

%\vspace{-.9cm}

\maketitle

\begin{abstract}
We study the problem of learning optimal policies in finite-horizon Markov Decision Processes (MDPs) using low-rank reinforcement learning (RL) methods. In finite-horizon MDPs, the policies, and therefore the value functions (VFs) are not stationary. This aggravates the challenges of high-dimensional MDPs, as they suffer from the curse of dimensionality and high sample complexity.
To address these issues, we propose modeling the VFs of finite-horizon MDPs as low-rank tensors, enabling a scalable representation that renders the problem of learning optimal policies tractable. 
{Our approach focuses on VF approximation within a policy iteration framework, where low-rank policy evaluation is combined with greedy policy improvement to compute near-optimal policies.}
We introduce an optimization-based framework for solving the Bellman equations with low-rank constraints, along with block-coordinate descent (BCD) and block-coordinate gradient descent (BCGD) algorithms, both with theoretical convergence guarantees.
{We further establish that bounded low-rank policy evaluation error translates into bounded policy improvement in the finite-horizon setting.}
For scenarios where the system dynamics are unknown, we adapt the proposed BCGD method to estimate the VFs using sampled trajectories.
Numerical experiments further demonstrate that the proposed framework reduces computational demands in controlled synthetic scenarios and more realistic resource allocation problems, {while achieving competitive policy performance in terms of attained returns.}
\end{abstract}


%%%%%%%%%%%%%%%%%%%%%%%%%%%%%%%%%%%%%%%%%%
\section{Introduction}
\label{S:intro}

Reinforcement learning (RL) has emerged as the leading framework for solving Markov Decision Processes (MDPs)\cite{bertsekas2012dynamic, puterman2014markov} in scenarios where system dynamics are unknown\cite{sutton2018reinforcement, bertsekas2019reinforcement}. Its recent advancements have powered groundbreaking achievements like AlphaGo~\cite{silver2016mastering, silver2017mastering} or ChatGPT~\cite{brown2020language}. Despite the predominant focus on infinite-horizon MDPs in the RL literature, many practical applications are characterized by finite decision-making horizons. Addressing this gap, the present work tackles the problem of learning optimal policies in finite-horizon MDPs via low-rank tensor decompositions.

RL is rooted in dynamic programming (DP), where the key challenge involves solving the Bellman equations (BEqs)~\cite{bellman1966dynamic}. This enables the estimation of the optimal value functions (VFs), which represent cumulative rewards. Optimal VFs are subsequently used to derive optimal policies, formalized as mappings that take a state as input and output an action. 
{In this work, we adopt this classical policy iteration perspective and focus on VF approximation as the core computational step for learning near-optimal policies in finite-horizon MDPs.}
The main difference between DP and RL is that DP is \emph{model-based}, relying on known MDP dynamics or a model, whereas RL is \emph{model-free}, using sampled data and stochastic approximation to estimate the VFs.
RL faces two primary challenges: (i) the curse of dimensionality, where the number of VFs to estimate grows exponentially with the size of the state and action spaces~\cite{sutton2018reinforcement, bertsekas2019reinforcement}, and (ii) high sample complexity, as RL methods often require a large number of samples to accurately estimate VFs~\cite{kakade2003sample}.
These challenges are further exacerbated in finite-horizon MDPs, where policies, and consequently VFs, are not stationary~\cite{puterman2014markov}. Even if the state happens to be the same at two different time points, the optimal action is likely to differ.
To alleviate these problems in infinite-horizon MDPs, VF approximation schemes have been proposed~\cite{bertsekas1996neuro}. These typically involve postulating \emph{parametric} mappings from state-action pairs to VFs, e.g., linear models or neural networks (NNs)~\cite{lagoudakis2003least, mnih2015human, cervino2021multi}. However, VF approximation has not been thoroughly studied in finite-horizon MDPs.

\vspace{.15cm}
\noindent \textbf{Contribution.}
This paper introduces \emph{tensor low-rank} methods to approximate the optimal VFs of finite-horizon MDPs. Specifically, we propose collecting the VFs into a tensor, where we explicitly enforce a low-rank PARAFAC structure. Under this model, the number of parameters to estimate scales linearly with the size of the state and action spaces, mitigating the curse of dimensionality. 
To estimate the tensor VFs, we propose an optimization-based formulation grounded in the BEqs. We propose two methods for solving this problem: a \emph{block-coordinate descent (BCD)} algorithm {(which corresponds to Alternating Least Squares (ALS) applied to a Bellman least-squares (LS) objective)} and a \emph{block-coordinate gradient descent (BCGD)} algorithm, both with convergence guarantees. When the MDP model is not available, we adapt the BCGD method to learn from sampled data, also with convergence guarantees. 
Furthermore, we connect the proposed stochastic approach to temporal difference (TD) learning~\cite{sutton2018reinforcement}, arguably the most celebrated framework for estimating VFs in RL.
{Importantly, we show that bounded low-rank evaluation error implies bounded policy improvement in the finite-horizon setting, thereby linking value approximation with policy performance guarantees.}
Numerical experiments confirm that our methods reduce the sample complexity needed for estimating the VFs accurately, {while achieving competitive performance in terms of return.}

The main contributions of this paper are:

\begin{enumerate}[left= 1pt .. 12pt, noitemsep]
    \item
    We propose modeling the VFs of finite-horizon MDPs using \emph{low-rank tensors}, providing an efficient approach to mitigate the curse of dimensionality.
    \item
    We introduce an \emph{optimization-based} approach rooted in the BEqs using \emph{low-rank tensors} and propose \emph{BCD} and \emph{BCGD} methods with guaranteed convergence.
    \item
    We adapt the proposed BCGD algorithm to learn \emph{stochastically} when the MDP model is unknown, and we also provide convergence guarantees. Furthermore, we connect the proposed algorithm with TD learning.
    \item
    We show numerically in two resource-allocation problems that low-rank tensors reduce computational demands in modeling VFs for finite-horizon MDPs.
\end{enumerate}

A preliminary version of this work was presented in a conference paper~\cite{rozada2024tensorb}. This journal version extends it by formalizing the optimization-based approach and connecting it to the BEqs. Additionally, we propose various convergent algorithms, provide a theoretical convergence analysis, and include a broader set of empirical experiments.

\vspace{.15cm}
\noindent \textbf{Prior work I: VF estimation in finite-horizon MDPs.} 
In finite-horizon setups, VFs are often estimated by reformulating the BEqs as a linear program~\cite{li2005lazy, kumar2015finite, bhattacharya2017linear, kalagarla2021sample}. However, the scalability of this approach is limited, making VF approximation the preferred strategy.
VF approximation has been extensively studied in infinite-horizon MDPs~\cite{bertsekas1996neuro, geist2013algorithmic, mnih2015human, sutton2018reinforcement}, but remains less explored for finite-horizon MDPs, particularly in the absence of the MDP model.
In the context of optimal control, NN-based approaches have been proposed in model-based problems to approximate VFs in finite-horizon MDPs~\cite{zhao2014neural, guzey2016neural, hure2021deep}. 
In model-free setups, fixed-horizon methods have been introduced to stabilize the convergence of VF estimation algorithms. 
These methods approximate infinite-horizon MDPs using finite-horizon ones, leveraging linear and NN-based function approximation techniques~\cite{dann2015sample, de2020fixed}.

\vspace{.15cm}
\noindent \textbf{Prior work II: Low-rank methods for VF estimation.}
Low-rank optimization has been extensively studied in tensor approximation problems~\cite{kolda2009tensor, gandy2011tensor, sidiropoulos2017tensor}, yet its application to estimating VFs remains largely unexplored.
{
In RL, tensor low-rank methods have been primarily used in other components of the learning problem. 
For instance, they have been applied to policy representations~\cite{rozada2025multilinear}, as well as to the estimation of transition probabilities in Markov chains~\cite{kressner2014low, georg2023low, navarro2025low} and MDPs~\cite{azizzadenesheli2016reinforcement, ni2019maximum, ni2023learning}. 
When the underlying MDP is low-rank, meaning that the transition matrix admits a low-rank factorization, efficient state-action representations can be constructed, leading to improved sample efficiency~\cite{agarwal2020flambe, uehara2023representation, huang2023reinforcement, zhang2023provably}.
However, low-rank transition dynamics do not imply that the corresponding VFs are low-rank. Conversely, VFs may show low-rank structure even when the underlying MDP is not low-rank. Despite this, the explicit exploitation of low-rank structure in VFs has received limited attention, especially with tensors.
For VF estimation, low-rank tensor representations have mainly appeared in model-based optimal control settings, particularly for infinite-horizon MDPs with quadratic cost functions~\cite{gorodetsky2015efficient, gorodetsky2018high}. 
More generally, least-squares tensor regression has been proposed to approximate VFs in both infinite-horizon~\cite{dolgov2021tensor} and finite-horizon MDPs~\cite{oster2022approximating}, where the coefficients of a linear VF approximation are represented as a tensor. 
In model-free settings, low-rank matrix~\cite{yang2019harnessing, shah2020sample, rozada2021low, sam2023overcoming, stojanovic2023spectral} and tensor models~\cite{tsai2021tensor, rozada2024tensor} have recently been proposed to approximate VFs in infinite-horizon MDPs. However, to the best of our knowledge, no prior work has proposed low-rank tensor methods to estimate VFs in finite-horizon MDPs. This paper aims to bridge this gap.
}

\vspace{.15cm}
\noindent \textbf{Outline.} Section~\ref{S:preliminaries} introduces the fundamentals of finite-horizon MDPs. Section~\ref{S:method_exact} details the tensor low-rank approach for approximating the BEqs with known dynamics, while Section~\ref{S:method_stochastic} extends it to sample-based scenarios. Section~\ref{S:results} evaluates the proposed methods, and Section~\ref{S:conclusion} concludes with key insights.
%%%%%%%%%%%%%%%%%%%%%%%%%%%%%%%%%%%%%%%%%%

%%%%%%%%%%%%%%%%%%%%%%%%%%%%%%%%%%%%%%%%%%
\section{Preliminaries}
\label{S:preliminaries}

We start by outlining the notation and reviewing the basics of tensor rank decompositions. Vectors are denoted by bold lowercase symbols $\bbq$, matrices by bold uppercase $\bbQ$, and tensors by underlined bold uppercase $\tenbQ$. Scalar indices,  $i$, are used for single-dimensional indexing, while boldface column vectors, $\bbi = (i_1, \dots, i_D)$, are used for $D$-dimensional indexing.
Vectors are indexed as $\bbq(i)$, matrices as $\bbQ(\bbi)=\bbQ(i_1, i_2)$ for a two-dimensional index, and tensors as $\tenbQ(\bbi)=\tenbQ(i_1, \hdots, i_D)$ for $D$-dimensional indices. The colon operator is used for element collection, where $\bbq(:)$ represents all elements of a vector, and $\bbQ(i, :)$ denotes $i$-th row of a matrix. For a 3-dimensional tensor, $\tenbQ(i, :, :)$ represents the matrix slice at the $i$-th index of the first dimension. For higher-order tensors, similar conventions apply.

For the purposes of this paper, we briefly review the basics of tensor algebra. Formally, a tensor $\tenbQ \in \reals^{N_1 \times \hdots \times N_D}$ is a $D$-dimensional array indexed by $\bbi = (i_1, \hdots, i_D)$, where $i_d \in \{1, \hdots, N_d\}$.
The dimensions of a tensor are also referred to as modes. Unlike matrices, tensors have multiple definitions of rank. Here, we focus on the PARAFAC rank, widely used for parsimonious data representations. Specifically, the PARAFAC rank is the smallest number of rank-1 tensors that, when added to each other, recover the original tensor~\cite{kolda2009tensor, sidiropoulos2017tensor}.
A tensor $\tenbZ$ is rank-1 if it can be expressed as the outer product of $D$ vectors, $\tenbZ = \bbz_1 \circledcirc \hdots \circledcirc \bbz_D$, which implies $\tenbZ(\bbi) = \prod_{d=1}^D \bbz_d(i_d)$. A tensor $\tenbQ$ has rank $K$ if it can be decomposed as $\tenbQ = \sum_{k=1}^K \bbq_1^k \circledcirc \hdots \circledcirc \bbq_D^k$.
The PARAFAC decomposition~\cite{bro1997parafac} identifies the vectors (factors) $\{\bbq_d^k\}_{k=1}^{K}$ for $d\!=\!1,..., D$, typically collected into matrices $\bbQ_d = [\bbq_d^1,..., \bbq_d^K]$, and is denoted  as $\tenbQ =$ $[[\bbQ_1, ..., \bbQ_D]]$. Using the entry-wise definition of the outer product, the $\bbi\!=\!(i_1, ..., i_D)$ entry of a rank-$K$ tensor is given by
\begin{equation}
    \label{eq::parafac_model}
    \tenbQ(\bbi) = \sum_{k=1}^K \prod_{d=1}^D \bbq_d^k(i_d) = \sum_{k=1}^K \prod_{d=1}^D \bbQ_d(i_d, k),    
\end{equation}
Equation~\eqref{eq::parafac_model} reveals why the PARAFAC decomposition provides a parsimonious representation: while the original tensor contains $\prod_{d=1}^D N_d$ elements, a number that grows exponentially with $D$, the PARAFAC representation requires only $( \sum_{d=1}^D N_d) K$ elements, growing linearly with $D$. Finally, as matrices can be reshaped into vectors, tensors can be reshaped too. Common reshaping includes expressing a tensor as a tall matrix with columns corresponding to one dimension or as vectors.
To simplify ensuing expressions, we define $\bigodot_{d=1}^D \bbQ_d=\bbQ_1 \odot \hdots \odot \bbQ_D$, with $\odot$ denoting the Khatri-Rao product. The PARAFAC decomposition yields the matricization along the $d$-th mode as $\mat_d(\tenbQ)=(\bigodot_{i = 1 \neq d}^D \bbQ_i)\bbQ_d^\Tr$, and the vectorization as $\vvec(\tenbQ)=(\bigodot_{d=1}^D \bbQ_d) \bbone$, where $\bbone$ denotes a vector of all ones.

\vspace{.2cm}
\noindent \textbf{Finite-horizon MDPs.}
We consider a finite-horizon MDP, denoted by the tuple $(\ccalS, \ccalA, P, R, H)$. Here, $\ccalS$ and $\ccalA$ are discrete state-action spaces with dimensions $D_s$ and $D_a$; $P(\cdot\,|\,s, a)$ is a probability measure over $\ccalS$ parametrized by the state-action pairs $(s, a) \in \ccalS \times \ccalA$; $R$: $\ccalS \times \ccalA \mapsto \reals$ is a deterministic reward function; and $H$ is a time horizon.
In finite-horizon MDPs, we consider a discrete time-space of finite duration $\ccalH=\{1,...,H\}$, where each element $h$ is a time index.
In time-step $h$, the agent observes the state $s_h$, and selects an action $a_h$. The environment transitions into a new state $s_{h+1}$ according to $P$, and provides a reward $r_h=R(s_h, a_h)$. The interaction with the environment stops after the $H$-th action $a_H$ is taken and the $H$-th reward $r_H$ is obtained.
Note that we use the lower-case $r$ to denote a reward in the sample-based context, representing the output of the reward function $R_h(s, a)$.

\begin{remark}
    Finite-horizon MDPs can have non-stationary transition probabilities $P = \{P_h\}_{h=1}^H$ and non-stationary stochastic reward functions $R = \{R_h\}_{h=1}^H$. For simplicity, we assume stationary $P$ and stationary deterministic $R$, though the results extend to non-stationary and stochastic cases.
\end{remark}

The goal is to learn a policy $\pi$ that maximizes the expected cumulative reward $\mathbb{E}_\pi \left[\sum_{h=1}^H r_h\right]$. In finite-horizon MDPs, policies are not stationary~\cite{puterman2014markov}, as the optimal actions depend on the time index $h$.
Thus, the objective is to learn a set of policies $\pi = \{\pi_h\}_{h=1}^H$, where each $\pi_h: \ccalS \mapsto \ccalA$ maps states to actions deterministically. The optimality of $\pi$ is evaluated by the expected cumulative rewards over the horizon $H$.
We can now define the VFs. For a given policy $\pi$, we consider the set of VFs $Q^\pi = \{ Q^\pi_h \}_{h=1}^H$, where $Q^\pi_h(s, a)$ is the expected reward conditioned on state-action pair $(s, a)$ at time step $h$
\begin{equation}
    \notag
    Q^\pi_h(s, a) = \mathbb{E}_\pi \Big[ \sum_{h'=h}^H r_{h'} | s_h=s, a_h=a \Big].
\end{equation}
The optimality of the VFs holds element-wise; for any policy $\pi$, the optimal VFs satisfy $Q^\star_h(s, a) \geq Q^\pi_h(s, a)$ for all triplets $(s, a, h)$~\cite{puterman2014markov, bertsekas2012dynamic}. Furthermore, an optimal policy $\pi^\star$ can be obtained from the optimal VFs $Q^\star = \{Q^\star_h\}_{h=1}^H$ by greedily maximizing the VFs with respect to (w.r.t.) actions, $\pi^\star_h(s) = \argmax_a Q^\star_h(s, a)$. Thus, by first computing the optimal VFs $Q^\star$, we can derive an optimal policy $\pi^\star$.

As a cornerstone of the DP literature, the BEqs provide an iterative method to compute $Q^\star$. More specifically, the BEqs characterize the VFs $Q^\pi$ for a fixed policy $\pi$~\cite{bellman1966dynamic}. At each time step $h \in \ccalH$, the VF for a state-action pair $(s, a)$ satisfies
\begin{equation}\label{eq::bellman_equation}
    Q_h^\pi(s, a) = R_h(s, a) + \sum_{s' \in \ccalS} P(s' | s, a) Q_{h+1}^\pi(s', \pi_{h+1}(s')),
\end{equation}
with $Q_{H+1}^\pi(s, a) = 0$ for all $(s, a) \in \ccalS \times \ccalA$. This convention holds throughout the paper. The BEqs form a linear system of equations that can be solved recursively, starting from the final step and working backward to the initial step.
Solving~\eqref{eq::bellman_equation} to compute the VFs for a given $\pi$ is known as \emph{policy evaluation}. Greedy maximization of $Q^\pi$ w.r.t. actions produces a new policy $\pi' = \{\pi'_h\}_{h=1}^H$, where
\begin{equation}
    \label{eq::policy_improvement}
    \pi'_h(s) = \argmax_a Q^\pi_h(s, a).
\end{equation}
This step, known as \emph{policy improvement}, produces a policy $\pi'$ that is guaranteed to outperform $\pi$ in terms of the attained VFs~\cite{bertsekas2012dynamic}. If $\pi'_h(s, a) = \pi_h(s, a)$ for all $s$, $a$, and $h$, then $\pi = \pi^\star$ and $Q^\pi = Q^\star$. This process underpins \emph{policy iteration}, an iterative method that alternates between policy evaluation and policy improvement to compute the optimal VFs $Q^\star$ and policy $\pi^\star$.
Algorithm~\ref{alg::policy_iteration} outlines the policy iteration method, where $\texttt{PolicyEvaluation}$ solves \eqref{eq::bellman_equation}, and $\texttt{PolicyImprovement}$ is defined in \eqref{eq::policy_improvement}. For Algorithm~\ref{alg::policy_iteration}, the parameter $\epsilon_1$ is the stopping criterion evaluated using a specified $\texttt{Dist}$ function, and $\beta_1$ denotes the hyper-parameters of the specific method used to implement policy evaluation.

Moreover, note that \eqref{eq::bellman_equation} also holds for the optimal policy. Hence, at each time step $h$, the optimal VF for a state-action pair $(s, a)$ satisfies the optimal BEqs, a necessary condition for optimality
\begin{equation}\label{eq::bellman_opt_equation}
    Q_h^\star(s, a) = R_h(s, a) + \sum_{s' \in \ccalS} P(s' | s, a) \max_{a'} Q_{h+1}^\star(s', a'),
\end{equation}
with $Q_{H+1}^\star(s, a) = 0$. The optimal BEqs are a non-linear system, but can also be solved recursively.

While the BEqs provide a sounded framework for estimating VFs, they require a separate VF for each state-action-timestep triplet, which renders policy iteration untractable in high-dimensional spaces. To overcome this problem, we propose representing the VFs as a tensor and use low-rank approximations to reduce the degrees of freedom of the tensor. This, which is the core contribution of this paper, is discussed in Section~\ref{S:method_exact}.

\begin{algorithm}[ht]
\caption{Policy Iteration}
\label{alg::policy_iteration}
\begin{algorithmic}[1]
\Require Random initial policy \( \pi^{(1)} \); MDP model $P$ and $R$; hyper-parameters $\beta_1$; maximum number of iterations $N$; and stopping criterion $\epsilon_1$.

\For{$n = 1$ to $N$}
    \State $Q^{\pi^{(n)}} \gets \texttt{PolicyEvaluation}(\pi^{(n)}, P, R, \beta_1)$
    \State $\pi^{(n+1)} \gets \texttt{PolicyImprovement}(Q^{\pi^{(n)}})$
    \If{$\texttt{Dist}(\pi^{(n+1)}, \pi^{(n)}) \leq \epsilon_1$~~} \textbf{Break}
    \EndIf
\EndFor

\State \textbf{Return} \( \pi^{(n+1)} \) 
\end{algorithmic}
\end{algorithm}

%%%%%%%%%%%%%%%%%%%%%%%%%%%%%%%%%%%%%%%%%%

%%%%%%%%%%%%%%%%%%%%%%%%%%%%%%%%%%%%%%%%%%
\section{Policy iteration via low-rank tensors}
\label{S:method_exact}

This section presents our structured policy evaluation framework for finite-horizon MDPs. We formulate VF approximation under a low-rank tensor model in Section~\ref{SS:problem}, and show how the resulting Bellman-error minimization problem can be solved via ALS and BCGD methods in Sections~\ref{SS:bcd} and~\ref{SS:bcgd}. {The proposed evaluation routines serve as the core component within a policy iteration scheme, enabling the computation of near-optimal policies under low-rank assumptions (Section \ref{SS:policy_iteration_guarantee}).}

\subsection{Problem formulation}
\label{SS:problem}

{
We address policy evaluation in finite-horizon MDPs by representing the VF as a structured tensor and approximating it with a low-rank PARAFAC model.
In infinite-horizon problems, VFs are commonly represented as state–action matrices. In finite-horizon settings, we instead have a collection of matrices $\{Q_h\}_{h=1}^H$, one per time step. Stacking them along the temporal axis yields a third-order tensor indexed by $(s,a,h)$.
However, in many applications the state and action spaces themselves admit a product structure. For instance, a dynamical system with 3D position and velocity has a six-dimensional state representation. Motivated by prior work on tensorized RL~\cite{tsai2021tensor, gorodetsky2018high, dolgov2021tensor, rozada2024tensor}, we explicitly exploit this structure.
Formally, we assume
\[
\mathcal{S} = \mathcal{S}_1 \times \dots \times \mathcal{S}_{D_s},
\qquad
\mathcal{A} = \mathcal{A}_1 \times \dots \times \mathcal{A}_{D_a}.
\]
The joint state–action–time space becomes
\[
\mathcal{D} = \mathcal{S} \times \mathcal{A} \times \mathcal{H},
\]
which induces a tensor of order $D = D_s + D_a + 1$.
The tensor order therefore equals the total number of independent coordinates indexing the VF (state components, action components, and time).
We collect all finite-horizon VFs into a $D$-dimensional tensor
$
\tenbQ \in \mathbb{R}^{|\mathcal{D}_1| \times \cdots \times |\mathcal{D}_D|},
$
where each triplet $(s,a,h)$ corresponds to one entry of $\tenbQ$. We use the shorthand
$
\tenbQ(\bbi_s,\bbi_a,h) = Q_h(s,a),
$
with $\bbi_s=(i_1, \hdots, i_{D_s})$ and $\bbi_a=(i_{D_s}+1, \hdots, i_{D_s + D_a})$.

In high dimensions, direct estimation of all $|\mathcal{D}|$ entries is rarely feasible. We therefore approximate $\tenbQ$ using a PARAFAC decomposition of rank $K$ as 
$
\tenbQ \approx [[\ccalQ]],
$
where $\ccalQ=\{\bbQ_1,\dots,\bbQ_D\}$ are the factor matrices and we use $\ccalQ_s$ and $\ccalQ_a$ to denote the collections of state and action factors, respectively.
Under this model, the number of parameters reduces to
\(
(\sum_{d=1}^D |\mathcal{D}_d|)K,
\)
which scales linearly in the tensor order.
}

Given a policy $\pi$, instead of solving the Bellman equations directly, we minimize the Bellman error under the PARAFAC parameterization
\begin{align}
    \notag
    &\delta_\ccalQ^\pi(s, a, h) =R_h(s, a)\\
    & + \sum_{s' \in \ccalS} P(s' | s, a) \tenbQ(\bbi_{s'}, \bbi_{\pi_{h+1}(s')}, h+1) - \tenbQ(\bbi_s, \bbi_a, h),\label{E:belmman_error}
\end{align}
which leads to the loss
\begin{equation}
L^\pi(\ccalQ)
=
\frac{1}{H}
\sum_{h=1}^H
\sum_{s,a}
\delta_\ccalQ^\pi(s,a,h)^2.
\label{eq::problem}
\end{equation}
The dependence of $L$ and $\delta_\ccalQ$ on $\pi$ will be dropped whenever clear from the context. 
Note that global minimizers of \eqref{eq::problem} satisfy the Bellman equations provided the rank $K$ is not smaller than the rank of the true value tensor.
{
If the rank $K$ is chosen too small, a minimizer may not exist \cite{de2008tensor}. Nevertheless, as we show in the following subsection, we can still ensure convergence to local optima.}
The resulting optimization problem is non-convex due to multiplicative interactions between factors, but it exhibits a favorable multi-convex structure. This structure is exploited via alternating optimization, as described next.

\subsection{Block-coordinate methods (ALS)}
\label{SS:bcd}

The objective in~\eqref{eq::problem} is non-convex due to the multiplicative interaction between tensor factors. However, it is \emph{multi-convex}: if all factors $\{\bbQ_i\}_{i\neq d}$ are fixed, the problem becomes convex in the remaining factor $\bbQ_d$.
{
If one fixes all factors except $\bbQ_d$, minimizing $L(\ccalQ)$ w.r.t. $\bbQ_d$ reduces to an LS problem. Consequently, cyclically updating each factor while keeping the others fixed corresponds exactly to ALS} applied to the Bellman LS objective.

{
To use a compact notation, consider that the $|\mathcal{S}||\mathcal{A}|H$ $\delta_\ccalQ(s, a, h)$ terms in \eqref{eq::problem}  are collected into a vector and, accordingly, define the extended reward vector $\bar{\bbr}\in
\mathbb{R}^{|\mathcal{S}||\mathcal{A}|H}$ as $H$ copies of vector $\bbr$. Then, for any fixed mode $d$, the cost $L(\ccalQ)$ \eqref{eq::problem} can be alternatively written as
\begin{equation}
L(\ccalQ)
=
\frac{1}{H}
\|\bar{\bbr} + \bar{\bbC}_{d^{\setminus}} \bbq_d\|_2^2,
\label{eq::problem_fixed_least_squares}
\end{equation}
where $\bbq_d = \mathrm{vec}(\bbQ_d^\Tr)\in\reals^{N_dK}$ and $\bar{\bbC}_{d^{\setminus}}$ is the design matrix associated with the LS problem. Its construction, which follows directly from vectorizing the Bellman residuals under our low-rank assumption, is provided in Appendix~\ref{S:prop_least_squares} for completeness [cf. \eqref{eq::c_def}].
The following lemma formalizes the connection between the optimization of \eqref{eq::problem} and \eqref{eq::problem_fixed_least_squares}.

\begin{lemma}\label{prop::least_squares}
For any mode $d$, minimizing $L(\ccalQ)$ in \eqref{eq::problem} w.r.t. $\bbQ_d$ while keeping the remaining factors fixed reduces to the LS problem of minimizing \eqref{eq::problem_fixed_least_squares} w.r.t. $\bbq_d$.
\end{lemma}
}

This interpretation allows us to leverage classical BCD theory~\cite{bertsekas1999nonlinear} while carefully accounting for the structure induced by the Bellman equations. Given $\ccalQ=\{\bbQ_1, \hdots, \bbQ_D\}$ , block-coordinate methods iteratively optimize one factor $\bbQ_d = \text{unvec}(\bbq_d) \in \ccalQ$ at a time while keeping the others fixed. 
This yields a Block-Coordinate Policy Evaluation (BC-PE) procedure summarized in Algorithm \ref{alg::policy_evaluation_bc}, where $\texttt{Update}$ is the specific block-coordinate method employed, with hyper-parameters $\beta_2$, and $\epsilon_2$ a stopping criterion. The factor iterates are defined as
\begin{equation}
    \label{eq::q_factor_iterates}
    \ccalQ_d^{(m)}=\{\bbQ_1^{(m)}, \hdots, \bbQ_{d-1}^{(m)}, \bbQ_{d}^{(m - 1)}, \hdots, \bbQ_D^{(m-1)}\},    
\end{equation}
with $m$ being the iteration index. 
At the end of each iteration, we normalize the factors to ensure they share the same Frobenius norm, i.e.,  $\|\bbQ^{(m)}_1\|_F = ... = \|\bbQ^{(m)}_D\|_F$. This normalization does not affect the PARAFAC decomposition due to its scale-counter-scale ambiguity, {but prevents from degenerate behaviors \cite{de2008tensor}.}

Since the subproblems in \eqref{eq::problem_fixed_least_squares} are LS problems, their closed-form solutions are tractable and therefore we can apply cyclic ALS updates over all modes. Specifically, we implement \texttt{Update} in \ref{alg::policy_evaluation_bc} with $\beta_2 = \{ \emptyset \}$ using the update rule
\begin{equation}\label{eq::q_factor_iterates_rule}
\bbq_d^{(m)}
=
-
(\bar{\bbC}_{d^{\setminus}}^{(m)})^\dagger
\bar{\bbr},    
\end{equation}
followed by reshaping into $\bbQ_d^{(m)}$.
We refer to the resulting BC-PE with the update rule described in \eqref{eq::q_factor_iterates_rule} as BCD-PE.
%After each full sweep over $d=1,\dots,D$, we normalize the factors to resolve the scale ambiguity inherent in PARAFAC models. 
%The resulting procedure is summarized in Algorithm~\ref{alg::policy_evaluation_bc}.

\begin{algorithm}[ht]
\caption{Block-Coordinate Policy Evaluation (BC-PE)}
\label{alg::policy_evaluation_bc}
\begin{algorithmic}[1]
\Require Policy $\pi$; MDP model $P$ and $R$; hyper-parameters $\beta_2$; number of iterations $M$; and stopping criterion $\epsilon_2$.

\State Initialize random factors \( \ccalQ^{(0)} \)
\For{$m = 1$ to $M$}
    \For{$d = 1$ to $D$}
        \State $\bbQ^{(m)}_d \gets \texttt{Update}(\ccalQ_d^{(m)}, \pi, P, R, \beta_2)$
    \EndFor
    \State $\bbQ^{(m)}_d \gets \frac{\left(\prod_{d=1}^D \| \bbQ^{(m)}_d \|_F\right)^{\frac{1}{D}}}{\|\bbQ^{(m)}_d\|_F} \bbQ^{(m)}_d$ for all $d \in \ccalD$
    \State
    \If{$\sum_{d=1}^D \| \bbQ^{(m)}_d - \bbQ^{(m-1)}_d \|_F^2 \leq \epsilon_2$}~~\textbf{Break}
    \EndIf
\EndFor

\State \textbf{Return} \( \ccalQ^{(m)} \) 
\end{algorithmic}
\end{algorithm}

{
BC-PE serves as the policy evaluation step within finite-horizon policy iteration. Given a policy $\pi$, we run BC-PE to estimate its low-rank value tensor. 
The policy is then improved greedily using the updated VF, and the procedure is repeated until convergence. 
Specifically, one first runs Algorithm \ref{alg::policy_iteration} and then, in line 2 the \texttt{PolicyEvaluation} function is implemented by running Algorithm \ref{alg::policy_evaluation_bc}. The parameters $\beta_1$ in line 2 are set to $\beta_1 = \{M, \epsilon_2, \beta_2\}$, since those are the inputs required by Algorithm  \ref{alg::policy_evaluation_bc}. Finally, a natural choice for the function \texttt{Dist} in line 4 used to stop Algorithm \ref{alg::policy_iteration} is the Frobenious norm of the difference between two iterations. The version of Algorithm \ref{alg::policy_iteration} run using BCD-PE in line 2 will be referred to as BCD-PI.}

Although ALS is classical, its application to Bellman-error minimization under a  tensor parameterization requires additional care.
In particular, the design matrices $\bar{\bbC}_{d^{\setminus}}$ depend on transition dynamics and temporal coupling across stages.
To establish convergence, we require the following mild rank condition and assuming that stationary points of \eqref{eq::problem} exist.

\begin{assumption}[Regularity of factor matrices]
\label{ass::regularity}
For all iterations $m$ and modes $d$, the design matrix 
$\bar{\bbC}_{d^{\setminus}}^{(m)}$ has full column rank.
\end{assumption}

{
\begin{assumption}[Existence of stationary points]
\label{ass::stationary}
The set of stationary points of problem \eqref{eq::problem} is non-empty.
\end{assumption}
}

{
The convergence analysis assumes that the LS subproblems admit unique solutions, ensured by Ass.~\ref{ass::regularity}. 
This condition is generically satisfied whenever $|\mathcal{D}_d| \ge K$ and the factors are initialized with full column rank, as in ALS-type methods. 
Furthermore, the normalization step in Algorithm~\ref{alg::policy_evaluation_bc} prevents the scaling degeneracy of the PARAFAC decomposition that may arise when a global minimizer does not exist (e.g., when $K$ is underestimated). 
Under this setting, Ass.~\ref{ass::stationary} ensures that stationary points exist and the algorithm can converge to them. 
This is a mild assumption, since the loss in~\eqref{eq::problem} is smooth, non-convex, and typically has stationary points.
In practice, $K$ is treated as a hyperparameter and selected by validation.}
With this in mind, we now state the convergence result.

\begin{theorem}\label{thm::bcd_convergence}
Let Ass.~\ref{ass::regularity} and \ref{ass::stationary} hold and suppose the initial factors $\ccalQ^{(0)}$ are bounded. 
Then the sequence $\{\ccalQ^{(m)}\}_{m \ge 0}$ generated by BC-PE satisfies:
{\begin{enumerate}
    \item The objective sequence $\{L(\ccalQ^{(m)})\}$ is monotonically non-increasing.
    \item The iterates remain bounded.
    \item Every limit point of $\{\ccalQ^{(m)}\}$ is a stationary point of~\eqref{eq::problem}.
\end{enumerate}
In particular, the sequence converges to the set of stationary points of~\eqref{eq::problem}.}
\end{theorem} %\red{AGM: There may be a couple of issues with the proof. Furthermore, better convergence claims may be feasible. Read related discussion in my answer to the second comment of reviewer 2.}

The proof adapts classical BCD theory~\cite{xu2013block} to the structured Bellman LS objective and is provided in Appendix~\ref{S:app_bcd_conv}.

\subsection{Leveraging gradient descent}
\label{SS:bcgd}

{While BCD-PE performs exact block updates via LS (i.e., ALS steps), it may become computationally demanding as the MDP grows in size, since it requires computing the pseudo-inverse of $\bar{\bbC}_{d^{\setminus}}^{(m)}$, whose cost scales cubically with the block dimension.} 
To reduce this computational burden, we consider an inexact block update based on gradient descent. Specifically, instead of solving each LS subproblem exactly, we perform a gradient descent step of $L(\ccalQ)$ w.r.t. $\bbQ_d$ while keeping the remaining factors fixed. This yields a BCGD scheme.

Concretely, the \texttt{Update} step in Algorithm~\ref{alg::policy_evaluation_bc} is modified to use $\beta_2 = \{\alpha\}$, where $\alpha>0$ is a step size. The corresponding update rule is
\begin{equation}
    \label{eq::bcgd_rule}
    \bbq_d^{(m)} 
    =
    \bbq_d^{(m-1)} 
    -
    2\alpha
    (\bar{\bbC}_{d^{\setminus}}^{(m)})^\Tr
    \big(
        \bar \bbr 
        +
        \bar{\bbC}_{d^{\setminus}}^{(m)} 
        \bbq_d^{(m-1)}
    \big),
\end{equation}
where $\bar{\bbC}_{d^{\setminus}}^{(m)}$ is computed from the current factor iterates $\ccalQ_d^{(m)}$, and $\bbq_d^{(m)}=\vvec(\bbQ_d^{(m)})$.
We refer to BC-PE equipped with the update rule in~\eqref{eq::bcgd_rule} as BCGD-PE, and to the corresponding policy iteration scheme as BCGD-PI.

Under the same structural assumptions used in the BCD analysis, convergence to a stationary point can also be established for BCGD-PE, provided that the step size $\alpha$ is sufficiently small. The result is formalized below, with proof given in Appendix~\ref{S:app_bcgd_conv}.

\begin{theorem}\label{thm::bcgd_convergence}
{Let Ass. \ref{ass::stationary} hold}. Let $\alpha>0$ be small enough and suppose the initial factors $\ccalQ^{(0)}$ are bounded. Then the sequence $\{\ccalQ^{(m)}\}_{m \ge 0}$ generated by BCGD-PE satisfies:
{\begin{enumerate}
    \item The objective sequence $\{L(\ccalQ^{(m)})\}$ is monotonically non-increasing.
    \item The iterates remain bounded.
    \item Every limit point of $\{\ccalQ^{(m)}\}$ is a stationary point of~\eqref{eq::problem}.
\end{enumerate}
In particular, the sequence converges to the set of stationary points of~\eqref{eq::problem}.}
\end{theorem}

{
Together, BCD-PE (exact ALS updates) and BCGD-PE (inexact gradient-based updates) provide 
two convergent approaches for solving~\eqref{eq::problem} under a fixed policy $\pi$. 
It is also instructive to compare their per-iteration computational costs.
The dominant cost of each BCD-PE block update is the computation of the pseudoinverse 
of $\bar{\bbC}_{d^{\setminus}}^{(m)} \in \reals^{|\ccalS||\ccalA|H \times |\ccalD_d| K}$, 
which scales as $O\big(|\ccalS||\ccalA|H \cdot (|\ccalD_d| K)^2\big)$ per block and 
$O\big(D \cdot |\ccalS||\ccalA|H \cdot (\max_d |\ccalD_d| K)^2\big)$ per full sweep.
Each BCGD-PE block update replaces this pseudoinverse with a single matrix-vector product, 
reducing the per-block cost to $O\big(|\ccalS||\ccalA|H \cdot |\ccalD_d| K\big)$ and the 
per-sweep cost to $O\big(D \cdot |\ccalS||\ccalA|H \cdot \max_d |\ccalD_d| K\big)$.
In both cases, the cost scales \emph{linearly} in $D$, $H$, and each dimension $|D_d|$, 
in contrast to the $O\big(\prod_{d=1}^D |\ccalD_d|\big)$ storage and update cost of tabular methods. 
This linear scaling is a direct consequence of the PARAFAC parameterization and is the 
primary source of the efficiency gains observed in the experiments.
When the MDP model is unavailable, the stochastic extension of BCGD-PE introduced in 
Section~\ref{S:method_stochastic} reduces the per-iteration cost further: 
processing a single sampled transition brings the per-block update down to $O(DK)$, 
independent of the MDP size.

Both methods, however, require access to the transition model $P=\{P_h\}_{h=1}^H$ and 
reward functions $R = \{R_h\}_{h=1}^H$. In many practical settings only sampled 
trajectories are available. This motivates the stochastic extensions developed in 
Section~\ref{S:method_stochastic}, where we derive and analyze sample-based counterparts 
of the proposed algorithms.

\subsection{From low-rank policy evaluation to near-optimal policies}
\label{SS:policy_iteration_guarantee}

Having established convergence of BC-PE and BCGD-PE, we now assess how low-rank policy evaluation translates into policy improvement. At iteration $n$, we run BC-PE or BCGD-PE under policy $\pi^{(n)}$ to obtain rank-$K$ factors $\ccalQ^{(n)}$, which synthesize the approximate value tensor $\tenbQ^{(n)}$. The updated policy is then
\[
\pi_h^{(n+1)}(s) = \argmax_a\, \tenbQ^{(n)}(\bbi_s, \bbi_a, h).
\]

To quantify the improvement from $\pi^{(n)}$ to $\pi^{(n+1)}$, we introduce two gaps defined in terms of the \emph{exact} value function $Q_h^{\pi^{(n)}}$ (i.e., the solution to~\eqref{eq::bellman_equation} for $\pi=\pi^{(n)}$) and the greedy action $a_h^\star(s) = \argmax_a Q_h^{\pi^{(n)}}(s,a)$. The first is the \emph{greedy advantage} of the current policy,
\begin{align}
    g_h^{(n)}(s) = Q_h^{\pi^{(n)}}(s, a_h^\star(s)) - Q_h^{\pi^{(n)}}(s, \pi_h^{(n)}(s)),
\end{align}
which measures how much the current policy $\pi^{(n)}$ falls short of greedy optimality and is independent of the approximation algorithm. The second is the \emph{approximation error},
\begin{align}
    \varepsilon_h^{(n)}(s) = Q_h^{\pi^{(n)}}(s, a_h^\star(s)) - Q_h^{\pi^{(n)}}(s, \pi_h^{(n+1)}(s)),
\end{align}
which captures how much the updated policy $\pi^{(n+1)}$, derived from the low-rank approximation, deviates from the true greedy action. Their difference directly quantifies the per-step improvement:
\begin{align}
g_h^{(n)}(s) - \varepsilon_h^{(n)}(s) = Q_h^{\pi^{(n)}}(s,\pi_h^{(n+1)}(s))- Q_h^{\pi^{(n)}}(s,\pi_h^{(n)}(s)).
\end{align}

Bounding $\varepsilon_h^{(n)}(s)$ requires controlling the approximation quality of $\tenbQ^{(n)}$. To this end, let $\ccalF_K$ denote the set of rank-$K$ factor collections, and decompose the Bellman residual as
\begin{align}
\ell_h^{(n)}(\ccalQ^{(n)}) = \inf_{\ccalQ\in\ccalF_K}\ell_h^{(n)}(\ccalQ) + \eta_h^{(n)},
\end{align}
where $\ell_h^{(n)}(\ccalQ):=\sum_{s,a}\delta_\ccalQ^{\pi^{(n)}}(s,a,h)^2$ is the per-step Bellman residual~\eqref{E:belmman_error}. 
The infimum term captures the best achievable rank-$K$ approximation error with respect to the tensor of true VFs, which collects $Q_h^{\pi^{(n)}}$ over all state-action pairs and has rank $K^\star$, and $\eta_h^{(n)} \geq 0$ is the suboptimality gap due to not reaching that best approximation. Summing over $h$ and $s$ and bounding each term in this decomposition (see Appendix~\ref{SS:performance_bound} for details) yields the following result.

\begin{theorem}
\label{thm::perf_bound}
Let $P_1$ denote the distribution of the initial state and define $J(\pi):=\mathbb{E}_{s\sim P_1}[V_1^\pi(s)]$. Then, there exist finite constants $\kappa_0, \kappa_1 > 0$ such that
\begin{align}
    \notag
    &J(\pi^{(n+1)}) - J(\pi^{(n)}) \geq \\
    \notag
    &\sum_{h=1}^H \mathbb{E}_{s_h \sim P_h^{\pi^{(n+1)}}}\!\left[g_h^{(n)}(s_h)\right]
    -\kappa_1 \left[K^\star - K\right]_+ - \kappa_0 \sum_{h=1}^H \eta_h^{(n)}.
\end{align}
\end{theorem}

Theorem~\ref{thm::perf_bound} shows that the per-iteration improvement is the expected 
greedy advantage of $\pi^{(n)}$, penalized by two terms. The first is the \emph{modeling 
error} $\kappa_1[K^\star - K]_+$, which vanishes when $K \leq K^\star$. Here $\kappa_1 = 
\kappa_0 \beta_{K+1}$, where $\beta_{K+1}$ is the infinity norm of the $(K+1)$-th factor 
of the true VF tensor, so the penalty grows with the magnitude of the factors discarded 
when truncating to rank $K$. If the true VF tensor is well approximated at rank $K$ 
(i.e., $\beta_{K+1} \approx 0$), the modeling error becomes negligible regardless of the 
gap $K^\star - K$. The second penalty is the \emph{optimization error} $\kappa_0\sum_h 
\eta_h^{(n)}$, which vanishes when BC-PE or BCGD-PE reaches the best rank-$K$ 
approximation. When both errors are small, the improvement is essentially governed by 
how suboptimal the current policy is, as measured by the greedy advantage. This also 
provides a practical justification for the empirical low-rankness observed in 
Section~\ref{SS:res_exact}, where $\beta_{K+1}$ is seen to decay rapidly with $K$.}

%%%%%%%%%%%%%%%%%%%%%%%%%%%%%%%%%%%%%%%%%%

%%%%%%%%%%%%%%%%%%%%%%%%%%%%%%%%%%%%%%%%%%
\section{Stochastic policy iteration}
\label{S:method_stochastic}

This section shows how to approximate optimal VFs with low-rank tensors when the MDP model is not available. Section~\ref{SS:problem_stochastic} introduces the stochastic counterpart of \eqref{eq::problem}, while Sections~\ref{SS:sbcgd_method} and~\ref{SS:td_method} outline methods to solve it.

\subsection{Stochastic problem formulation}
\label{SS:problem_stochastic}

When the MDP model is unavailable, solving \eqref{eq::problem} directly is not feasible. Both BCD-PE and BCGD-PE rely on $R$ and $P$ to compute $\bar{\bbr}$ and $\bar{\bbC}_{d^{\setminus}}$. However, if we can sample trajectories from the MDP, we can reformulate \eqref{eq::problem} as a stochastic optimization problem. A natural approach to do so involves drawing trajectories and empirically averaging Bellman errors.
To formalize this, we denote a transition as $\sigma = (s, a, r, s', a', h)$, where $a$ and $a'$ are selected according to the policy $\pi$, i.e., $a = \pi_h(s)$ and $a' = \pi_{h+1}(s')$. The \emph{empirical Bellman error} is defined as
\begin{align}
    \notag
    &\tilde \delta_\ccalQ(\sigma) = r + \tenbQ(\bbi_{s'}, \bbi_{a'}, h+1) - \tenbQ(\bbi_s, \bbi_a, h)
\end{align}
where, as usual, $\tenbQ = [[\ccalQ]]$. A trajectory $\tau$ is just a collection of $H$ transitions from $h=1$ to $h=H$.
{
The transition dynamics $P$, the policy $\pi$, and the initial state distribution $P_1$ induce a probability measure over the space of trajectories $\ccalT$, which we denote by $\mu^\pi$. The probability mass function of $\mu^\pi$ for a trajectory $\tau = (s_1,\dots,s_H)$ is given by $\mu^\pi(\tau) = P_1(s_1)\prod_{h=1}^{H-1} P(s_{h+1}\mid s_h,\pi_h(s_h))$.
}
Using this, we define the following stochastic optimization
\begin{align}
    \label{eq::stochastic_problem_trajectories}
    &\min_\ccalQ  L_{\mu^\pi}(\ccalQ), \;\; \text{with} \; L_{\mu^\pi}(\ccalQ) \! = \! \mathbb{E}_{\tau \sim \mu^\pi} \!\! \left[\frac{1}{H} \!\! \sum_{\sigma_h \in \tau} \! \tilde \delta_\ccalQ(\sigma_h)^2 \right] \!\!.
\end{align}

Problem \eqref{eq::stochastic_problem_trajectories} is amenable to stochastic methods. If we can efficiently sample trajectories according to $\mu^\pi$, stochastic approximations can be constructed to solve it. However, processing entire trajectories to minimize \eqref{eq::stochastic_problem_trajectories} can be computationally prohibitive. To address this, RL typically evaluates errors using individual transitions, allowing for immediate updates of the VFs with each sampled transition. This approach is well-suited for online and incremental methods, which require sampling based on a probability measure over the transition space.
{
To define this probability measure, consider the probability measure over $\ccalS$ under policy $\pi$ at time-step $h$, denoted by $P^\pi_h(\cdot)$. We also define a probability distribution over $\ccalH$ as $P_\ccalH(h) = 1/H$. Combined, $P_\ccalH$, $P^\pi_h$, $P$, and $\pi$ induce a probability measure on the space of transitions, characterized by the probability mass function $\xi^\pi$, defined for a transition $\sigma=(s,a,s',h)$ as $\xi^\pi(\sigma)=P_\ccalH(h)\,P^\pi_h(s)\,P(s'\mid s,\pi_h(s))$. 
This yields
}
\begin{align}
    \label{eq::stochastic_problem_samples}
    &\min_\ccalQ \; L_{\xi^\pi}(\ccalQ), \quad \text{with} \; L_{\xi^\pi}(\ccalQ)= \mathbb{E}_{\sigma \sim \xi^\pi}  \left[ \tilde \delta_\ccalQ(\sigma)^2 \right],
\end{align}
which instead of relying on full trajectories, only requires transitions sampled from $\xi^\pi$ to do the updates.
In a nutshell, two stochastic approximations to \eqref{eq::problem} have been proposed: \eqref{eq::stochastic_problem_samples}, which relies only on individual samples, and \eqref{eq::stochastic_problem_trajectories}, which is a bit more demanding and learns from entire trajectories. 
Interestingly, \eqref{eq::stochastic_problem_trajectories} and \eqref{eq::stochastic_problem_samples} lead to the same optimal solution, as stated in the following theorem, whose proof is provided in Appendix~\ref{S:app_traj_samp_equ}.

\begin{theorem}
    \label{thm::traj_samp_equ}
    It holds that $L_{\mu^\pi}(\ccalQ) = L_{\xi^\pi}(\ccalQ)$. As a result, \eqref{eq::stochastic_problem_trajectories} and \eqref{eq::stochastic_problem_samples} are equivalent and their solutions $\ccalQ^\star_\mu$ and $\ccalQ^\star_\xi$ are equivalent as well.
\end{theorem}

\noindent {
Theorem~\ref{thm::traj_samp_equ} establishes the equivalence between \eqref{eq::stochastic_problem_trajectories} and \eqref{eq::stochastic_problem_samples}. 
The proof shows that the state distribution $P^\pi_h$ corresponds to the marginal distribution induced by policy $\pi$ at time-step $h$, obtained by averaging over all trajectories up to that time. Consequently, sampling a state $s_h$ from $P^\pi_h$ is equivalent to executing policy $\pi$ and following a trajectory $\tau$ until step $h$. This observation implies that trajectories generated by interacting with the environment naturally produce samples from the transition distribution, allowing updates to be performed after each observed transition while still optimizing the original trajectory-based objective in \eqref{eq::stochastic_problem_trajectories}.
}
%Thus, in the next section, we focus on solving \eqref{eq::stochastic_problem_samples}.

\begin{figure*}[t]
    \begin{center}
    \includegraphics[width=\linewidth]{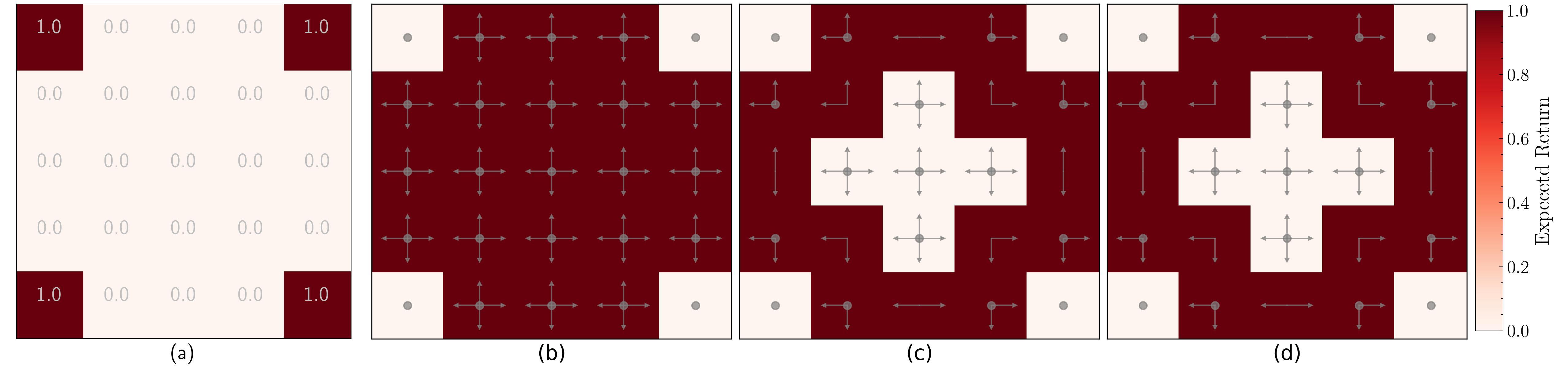}
    \vspace{-.3cm}
    \caption{Picture (a) shows the rewards of the grid-world environment. The remaining pictures show in time-step $3$ the estimated VFs and policy by (b) infinite-horizon policy iteration, (c) finite-horizon policy iteration, and (d) \textbf{BCD-PI}. In finite-horizon settings, the VFs and the policy are time-dependent.}
    \label{fig::grid_setup}
    \end{center}
    \vspace{-.6cm}
\end{figure*}

\subsection{Stochastic block coordinate (BC) methods}
\label{SS:sbcgd_method}

We propose stochastic BC methods to solve \eqref{eq::stochastic_problem_samples}, extending the principles of BC-PE to stochastic setups. The key idea is to iteratively update each factor in $\ccalQ$ while keeping the others fixed, using stochastic samples of the MDP. It is crucial to consider the Markovian nature of the sampling process, as it can bias the updates.
Specifically, we sample a trajectory from the MDP. At iteration $m$, we observe the transition $\sigma_m$ and we store it in an experience replay (ER) buffer $\ccalB$. 
Next, we sample a transition $\hat \sigma_m$ from the ER buffer $\ccalB$ uniformly at random and update every factor $\bbQ_d \in \ccalQ$ cyclically. The use of the ER buffer decorrelates the sampling process, mitigating the bias introduced by the Markov sampling \cite{fedus2020revisiting}. 
This procedure is summarized in the stochastic BC-PE algorithm shown in Algorithm \ref{alg::policy_evaluation_sbc}, where \texttt{StochUpdate} refers to the stochastic BC method with hyper-parameters $\beta_3$, and $\epsilon_3$ is the stopping criterion. The iterates $\ccalQ_d^{(m)}$ are defined in \eqref{eq::q_factor_iterates}.
As with BC-PE, the stochastic BC-PE can be used to estimate optimal policies when combined with policy iteration. In this case, we run Algorithm \ref{alg::policy_iteration} and then, when reaching line 2, we implement the function \texttt{PolicyEvaluation} using Algorithm \ref{alg::policy_evaluation_sbc} and setting parameters $\beta_1$ to $\beta_1 = \{M, \epsilon_3, \beta_3\}$, which are the inputs required by Algorithm \ref{alg::policy_evaluation_sbc}.

\begin{algorithm}[ht]
\caption{Stochastic BC-PE}
\label{alg::policy_evaluation_sbc}
\begin{algorithmic}[1]
\Require Policy $\pi$; MDP model $P$ and $R$; hyper-parameters $\beta_3$; number of iterations $M$; and stopping criterion $\epsilon_3$.

\State Initialize random factors \( \ccalQ^{(0)} \), empty buffer $\ccalB$ and $m=1$
\While{true}
    \State Sample initial state $s_m$
    \For{$h_m = 1$ to $H$}
        \State Sample $s'_m, r_m$ following $P$, $R$ and $a = \pi_{h_m}(s_m)$
        \State Set $a'_m = \pi_{h_m+1}(s'_m)$ and $s_{m+1} = s'_m$ 
        \State $\sigma_m \gets (s_m, a_m, r_m, s'_m, a'_m, h_m)$
        \State Store $\sigma_m$ in buffer $\ccalB$ and sample $\hat \sigma_m \sim \ccalB$
        \For{$d = 1$ to $D$}
            \State $\bbQ^{(m)}_d \gets \texttt{StochUpdate}(\ccalQ_d^{(m)}, \pi, \hat \sigma_m, \beta_3)$
        \EndFor
        \State $\bbQ^{(m)}_d \gets \frac{\left(\prod_{d=1}^D \| \bbQ^{(m)}_d \|_F\right)^{\frac{1}{D}}}{\|\bbQ^{(m)}_d\|_F} \bbQ^{(m)}_d$ for all $d \in \ccalD$
        \If{$\sum_{d=1}^D \| \bbQ^{(m)}_d - \bbQ^{(m-1)}_d \|_F^2 \leq \epsilon_3$~~} \textbf{Break}
        \EndIf
        \State $m \gets m + 1$
    \EndFor
\EndWhile

\State \textbf{Return} \( \ccalQ^{(m)} \) 
\end{algorithmic}
\end{algorithm}

For the specific update used in line 12 of Algorithm \ref{alg::policy_evaluation_sbc}, we propose extending BCGD-PE to the stochastic setting, by using stochastic gradients of \eqref{eq::stochastic_problem_samples}. With that in mind, for the transition $\hat \sigma_m=(\hat s_m, \hat a_m, \hat r_m, \hat s'_m, \hat a'_m, \hat h_m)$ we introduce the multi-dimensional indices $\bbi_m$ and $\bbi'_m$ associated with the triplets $(\hat  s_m, \hat a_m, \hat h_m)$ and $(\hat s'_m, \hat a'_m, \hat h_m+1)$, respectively. Next, leveraging the chain rule, we compute the stochastic gradients
\begin{align}
     \notag
    \tilde \nabla_{\bbQ_d} L_{\xi^\pi}\!(\ccalQ) \!=\! \nabla_{\bbQ_d} \tilde \delta_\ccalQ(\hat \sigma_m)^2\! = 2 \tilde \delta_\ccalQ(\hat \sigma_m) \nabla_{\bbQ_d} \!(\tenbQ(\bbi'_m) -  \tenbQ(\bbi_m) ) \\
    = 2 \tilde \delta_\ccalQ(\hat \sigma_m) \!\left(\!\! \bbe_{\bbi'_m(d)}\!\!\!\!\bigodot_{j=1\neq d}^D
    \!\!\!\bbQ_j(\bbi'_m(j), :) - \bbe_{\bbi_m(d)}\! \!\!\!\bigodot_{j=1\neq d}^D \!\!\!\bbQ_j(\bbi_m(j), :)\!\!\right)\!,   \label{eq::stochastic_gradient}
\end{align}
where $\bbe_{\bbi_m(d)} \in \{0, 1\}^{|\ccalD_d| \times 1}$ is a selection column vector with a single $1$ in the $\bbi_m(d)$-th entry.
The implementation of \texttt{StochUpdate} in Algorithm \ref{alg::policy_evaluation_sbc} becomes a stochastic gradient. Setting $\beta_3 = \{\{ \alpha^{(m)} \}^M_{m=1} \}$, where $\alpha^{(m)}$ is the step size, yields
\begin{equation}
    \label{eq::stochastic_bcgd_rule}
    \bbQ_d^{(m)} = \bbQ_d^{(m-1)} - \alpha^{(m)} \nabla_{\bbQ_d} \tilde \delta_{\ccalQ_d^{(m)}}(\hat \sigma_m)^2,
\end{equation}
where $\ccalQ_d^{(m)}$ is defined in \eqref{eq::q_factor_iterates} and $\nabla_{\bbQ_d} \tilde \delta_{\ccalQ_d^{(m)}}(\hat \sigma_m)^2$ is computed using \eqref{eq::stochastic_gradient}. We refer to the stochastic BC-PE algorithm with the update rule in \eqref{eq::stochastic_bcgd_rule} as S-BCGD-PE, and the corresponding policy iteration algorithm as S-BCGD-PI.
S-BCGD-PE converges to a stationary point of \eqref{eq::stochastic_problem_samples}, provided it does not generate unbounded iterates, which is formalized in the following assumption.

\begin{assumption}
    \label{ass::bounded_iterates}
    S-BCGD-PE generates bounded iterates, i.e. $\limsup_{m \to \infty} \| \bbQ_d^{(m)} \|_F < \infty$ for all $d=1, \hdots, D$.
\end{assumption}

\noindent The convergence result is formalized in the following theorem, whose proof is provided in Appendix~\ref{S:app_sbcgd_conv}.

\begin{theorem}\label{thm::sbcgd_convergence}
    Let Ass.  \ref{ass::bounded_iterates} hold, and select stepsizes $\{ \alpha^{(m)} \}_{m=1}^\infty$ such that 
    \begin{equation}
        \notag
        \sum_{m=1}^\infty \alpha^{(m)} = \infty \; \; \text{and} \;\; \sum_{m=1}^\infty (\alpha^{(m)})^2 < \infty.
    \end{equation}
    Then, for any initial bounded factors $\ccalQ^{(0)}$, solving~\eqref{eq::stochastic_problem_samples} via S-BCGD-PE generates a sequence $\{\ccalQ^{(m)}\}_{m \geq 1}$ that converges in expectation, i.e. $\lim_{m \to \infty} \mathbb{E} \left[ \| \nabla L_{\xi^\pi} (\ccalQ^{(m)}) \| \right]=0$.
\end{theorem}

%S-BCGD-PE provides a convergent approach for solving \eqref{eq::stochastic_problem_samples}, yet it presents a drawback. The empirical Bellman error, and therefore its gradient, are biased estimates, i.e. $\mathbb{E}[\tilde{\delta}_\ccalQ(\hat \sigma)^2] \neq \delta_\ccalQ(\hat \sigma)^2$. This issue is well-known in the RL literature \cite{geist2013algorithmic}, but it can be addressed via TD methods.

S-BCGD-PE provides a convergent approach to solving \eqref{eq::stochastic_problem_samples} but has a drawback; the empirical Bellman error and its gradient are biased estimates, i.e., $\mathbb{E}[\tilde{\delta}_\ccalQ(\hat \sigma)^2] \neq \delta_\ccalQ(\hat \sigma)^2$. This issue is well-known in RL \cite{geist2013algorithmic} and can be mitigated using TD methods.

\subsection{Temporal-difference methods}
\label{SS:td_method}

The standard approach in RL for an unbiased stochastic formulation is TD learning. In essence, given a transition $\sigma = (s, a, r, s', a', h)$, TD updates compute the gradient only w.r.t. the parameters of the VF model involved in the current step, i.e., those dependent on the pair $(s, a)$. 
Formally, at iteration $m$ we consider the factors from the previous iteration $\ccalQ^{(m-1)}$, and optimize over $\tenbQ$. The TD update uses the following empirical Bellman error
\begin{align}
    \notag
    &\tilde \delta_{\ccalQ, \ccalQ^{(m-1)}}(\sigma) = r + \tenbQ^{(m-1)}(\bbi_{s'}, \bbi_{a'}, h+1) - \tenbQ(\bbi_s, \bbi_a, h),
\end{align}
which approximates
\begin{align}
    \notag
    &\delta_{\ccalQ, \ccalQ^{(m-1)}}(s, a, h) = R_h(s, a)\\
    \notag
    &+ \sum_{s' \in \ccalS} P(s' | s, a) \tenbQ^{(m-1)}(\bbi_{s'}, \bbi_{a'}, h+1) \! - \! \tenbQ(\bbi_s, \bbi_a, h),
\end{align}
where $\tenbQ^{(m-1)} = [[\ccalQ^{(m-1)}]]$, $\tenbQ = [[\ccalQ]]$, and $a'=\pi_{h+1}(s')$. Then, TD solves the following iterative optimization
\begin{align}
    \label{eq::stochastic_problem_samples_td}
    \ccalQ^{(m)} = &\argmin_\ccalQ  L^{(m)}_{\xi^\pi}(\ccalQ),\\ 
    \notag
    &\text{with} \;\; L^{(m)}_{\xi^\pi}(\ccalQ) = \mathbb{E}_{\sigma \sim \xi^\pi} \left[ \tilde \delta_{\ccalQ, \ccalQ^{(m-1)}}(\sigma)^2 \right].
\end{align}
Clearly, solving \eqref{eq::stochastic_problem_samples_td} with $\ccalQ^{(m-1)} = \ccalQ^{(m)}$ provides a solution to the BEqs.  
However, note that the biasedness issue of the empirical Bellman errors remains, as $\mathbb{E}[\tilde{\delta}_{\ccalQ, \ccalQ^{(m-1)}}(\sigma)^2] \neq \delta_{\ccalQ, \ccalQ^{(m-1)}}(\sigma)^2$.  
Nevertheless, the stochastic gradient of $L^{(m)}_{\xi^\pi}(\ccalQ)$ w.r.t. $\bbQ_d$, which is given by  
\begin{align}
    \label{eq::stochastic_gradient_td}
    \tilde \nabla_{\bbQ_d} L^{(m)}_{\xi^\pi}(\ccalQ) &= \nabla_{\bbQ_d} \tilde \delta_{\ccalQ, \ccalQ^{(m-1)}}(\sigma_m)^2 \\
    \notag
    &= -2 \tilde \delta_{\ccalQ, \ccalQ^{(m-1)}}(\sigma_m) \nabla_{\bbQ_d} \tenbQ(\bbi_m)  \\
    \notag
    &= -2 \tilde \delta_{\ccalQ, \ccalQ^{(m-1)}}(\sigma_m) \bbe_{\bbi_m(d)} \!\!\! \bigodot_{j=1\neq d}^D \bbQ_j(\bbi_m(j), :),
\end{align}
ensures $\mathbb{E}[\tilde \nabla_{\bbQ_d} \tilde{\delta}_{\ccalQ, \ccalQ^{(m-1)}}(\sigma)^2] = \nabla_{\bbQ_d} \mathbb{E}  \left[ \delta_{\ccalQ, \ccalQ^{(m-1)}}(\sigma)^2 \right]$, so that the descent direction remains unbiased. This motivates the use of the stochastic gradient in \eqref{eq::stochastic_gradient_td} to develop a stochastic BC TD method, which mimics S-BCGD-PE but replaces \eqref{eq::stochastic_gradient} with \eqref{eq::stochastic_gradient_td}. This method is implemented by running Algorithm \ref{alg::policy_evaluation_sbc} and, when reaching line 12, setting $\beta_3 = {\{ \alpha^{(m)} \}_{m=1}^M}$ and using as stochastic update
\begin{equation}
    \label{eq::stochastic_td_rule}
    \bbQ_d^{(m)} = \bbQ_d^{(m-1)} - \alpha^{(m)}\nabla_{\bbQ_d} \tilde \delta_{\ccalQ_d^{(m)}, \ccalQ^{(m-1)}}(\hat \sigma_m)^2,
\end{equation}
where $\hat \sigma_m$ has been sampled from $\ccalB$. We refer to the stochastic BC-PE algorithm with the update rule in \eqref{eq::stochastic_td_rule} as BCTD-PE, and the corresponding policy iteration algorithm as BCTD-PI.

\begin{remark}
    Note that with minor modifications, setting $M=1$ in BCTD-PE implies that at iteration $m$, the policy $\pi$ is determined by the factors from the previous iteration $\ccalQ^{(m-1)}$, i.e. $a' = \argmax_{a} \tenbQ^{(m-1)}(\bbi_{s'}, \bbi_{a}, h+1)$. Consequently, the empirical Bellman error becomes
    \[
        r_m + \max_{a} \tenbQ^{(m-1)}(\bbi_{s'_m}, \bbi_{a}, h_m+1) - \tenbQ(\bbi_{s_m}, \bbi_{a_m}, h_m),
    \]
    which can be interpreted as the empirical error of the optimal BEqs \eqref{eq::bellman_opt_equation}. This is the stochastic approximation of the value iteration method using low-rank tensors, and is the main contribution of the conference version of this paper \cite{rozada2024tensorb}.
\end{remark}

TD methods, however, introduce an alternative limitation. While the formulation in \eqref{eq::stochastic_problem_samples_td} is motivated by the BEqs, it is possible for \( L_{\xi^\pi}^{(m)}(\ccalQ^{(m)}) = 0 \) to hold despite \( \ccalQ^{(m)} \neq \ccalQ^{(m-1)} \). This inconsistency violates the Bellman structure, as the successive factor iterates should align to satisfy the BEqs.
Furthermore, convergence of TD methods has been a long-standing challenge in RL \cite{bhandari2018finite, asadi2024td}.
In a nutshell, we proposed two alternative approaches for policy evaluation: S-BCGD-PE, which is biased but convergent, and BCTD-PE, which is unbiased yet not convergent. Both methods are run in combination with policy improvement to define S-BCGD-PI and BCTD-PI, that are used to estimate optimal policies.

%%%%%%%%%%%%%%%%%%%%%%%%%%%%%%%%%%%%%%%%%%

\begin{figure*}[t]
    \begin{center}
    \includegraphics[width=\linewidth]{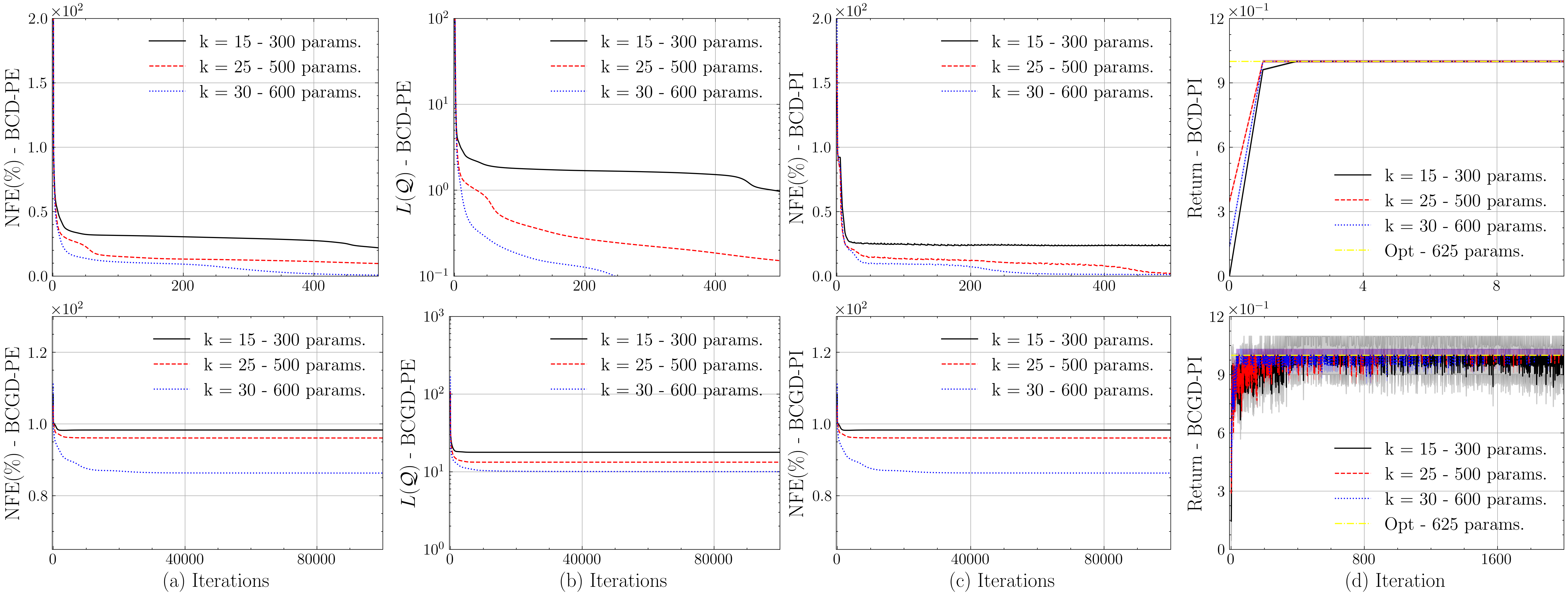}
    \vspace{-.3cm}
    \caption{The figure shows results for \textbf{BCD-PE} and \textbf{BCD-PI} in the first row, and \textbf{BCGD-PE} and \textbf{BCGD-PI} in the second. The columns display: (i) PE convergence in terms of (a) NFE and (b) $L(\ccalQ)$; and (ii) PI convergence in terms of (c) NFE and (d) empirical return.}
    \label{fig::grid_res}
    \end{center}
    \vspace{-.6cm}
\end{figure*}

\begin{figure}
    \centering
    \includegraphics[width=0.75\linewidth]{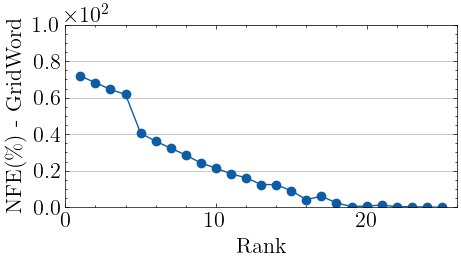}
    \caption{$\mathrm{NFE}$ between the tensor of optimal VFs obtained via policy iteration low-rank PARAFAC decomposition in the grid-like setup.}
    \label{fig::low_rank}
    \vspace{-.2cm}
\end{figure}

%%%%%%%%%%%%%%%%%%%%%%%%%%%%%%%%%%%%%%%%%%
\section{Numerical Analysis}
\label{S:results}

This section presents an evaluation of the proposed algorithms across various finite-horizon MDPs. The goal is to show that tensor low-rank methods outperform alternative approaches in terms of computational efficiency and sample complexity.
Section \ref{SS:res_exact} focuses on testing \textbf{BCD-PE} and \textbf{BCGD-PE} in a grid-world setup with a known MDP model.
{
Section \ref{SS:res_gym} evaluates \textbf{S-BCGD-PI} and \textbf{BCTD-PI} on two well-known control problems from the Gymnasium benchmark \cite{towers2024gymnasium}, 
whereas Section \ref{SS:res_stochastic} considers three realistic resource allocation problems with unknown MDP models.
}
The implementation details and the code are available in the repository \cite{rozada2024code}.

\subsection{Policy iteration with known MDP}
\label{SS:res_exact}

This subsection evaluates the performance of \textbf{BCD-PE} and \textbf{BCGD-PE} in a controlled grid-world setup with a known model, providing insights into the theoretical convergence guarantees from Theorems \ref{thm::bcd_convergence} and \ref{thm::bcgd_convergence}.
The environment is a $5 \times 5$ grid which illustrates the challenges of finite-horizon MDPs. The agent moves in four directions (up, right, left, down) or stays in the same cell. The horizon is $H=5$ steps, with rewards of $1$ placed in each corner (see Fig. \ref{fig::grid_setup}-a). The dynamics are deterministic.
If the horizon is infinite, all state-action pairs have equal value because any corner can be reached without time constraints, and there is no penalty for movement (Fig. \ref{fig::grid_setup}-b). In contrast, the finite-horizon setting prioritizes movements toward corners that are reachable within the remaining steps (Figs. \ref{fig::grid_setup}-c and \ref{fig::grid_setup}-d).

\vspace{.1cm}
\noindent \textbf{Empirical Low-Rankness.} We empirically verified the low-rank structure of the tensor representing the VFs, a 4-dimensional tensor with two dimensions for the state space, one for the action space, and one for time. The tensor size is $5 \times 5 \times 5 \times 5$ with 625 parameters and a maximum possible rank of $K_\text{max} = 125$.
The optimal VFs were computed by solving the BEqs via policy iteration, and the tensor was then approximated using PARAFAC decomposition for various ranks. Approximation performance was measured using the Normalized Frobenius Error (NFE), defined as $\mathrm{NFE} = ||\hat \tenbQ - \tenbQ||_F / ||\tenbQ||_F$.
As shown in Fig.~\ref{fig::low_rank}, $\mathrm{NFE}$ decreases sharply with increasing rank, indicating that most of the energy is captured by the first few factors. With a rank of $20$, the tensor achieves a negligible $\mathrm{NFE}$, closely approximating the optimal VFs and demonstrating the effectiveness of low-rank approximations. These results align with \cite{rozada2024tensor}, which also report low-rank structures in VFs of classical infinite-horizon MDPs when organized as tensors.

\vspace{.1cm}

\begin{figure}[t]
    \begin{center}
    \includegraphics[width=0.8\linewidth]{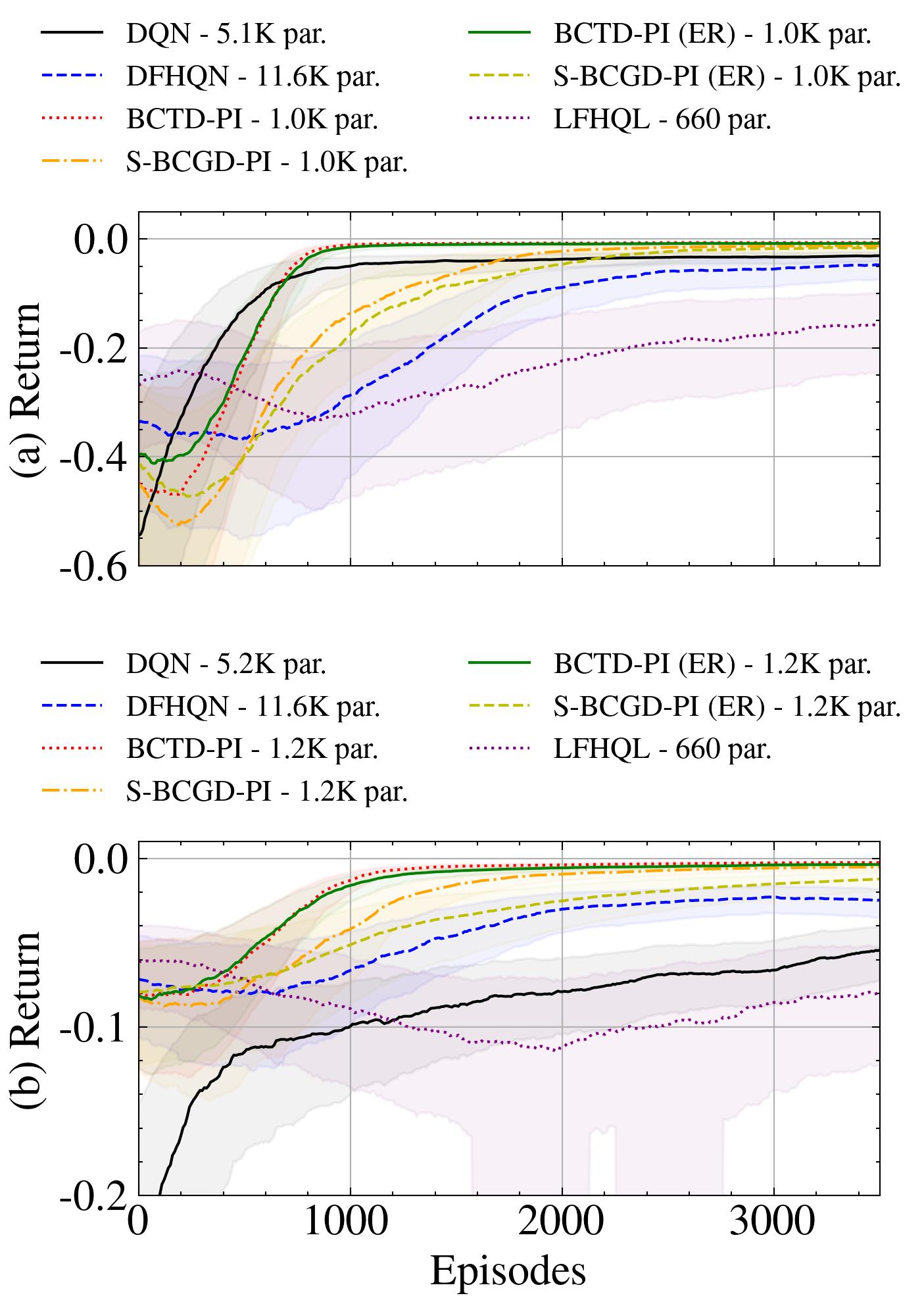}
    \vspace{-.4cm}
    \caption{Average return for \textbf{S-BCGD-PI} and \textbf{BCTD-PI} against different baselines for (a) the Pendulum and for (b) CartPole problems. \textbf{S-BCGD-PI} and \textbf{BCTD-PI} converge faster while requiring less number of parameters.}
    \label{fig::gym_res}
    \end{center}
    \vspace{-.6cm}
\end{figure}

\begin{figure}[t]
    \begin{center}
    \includegraphics[width=0.8\linewidth]{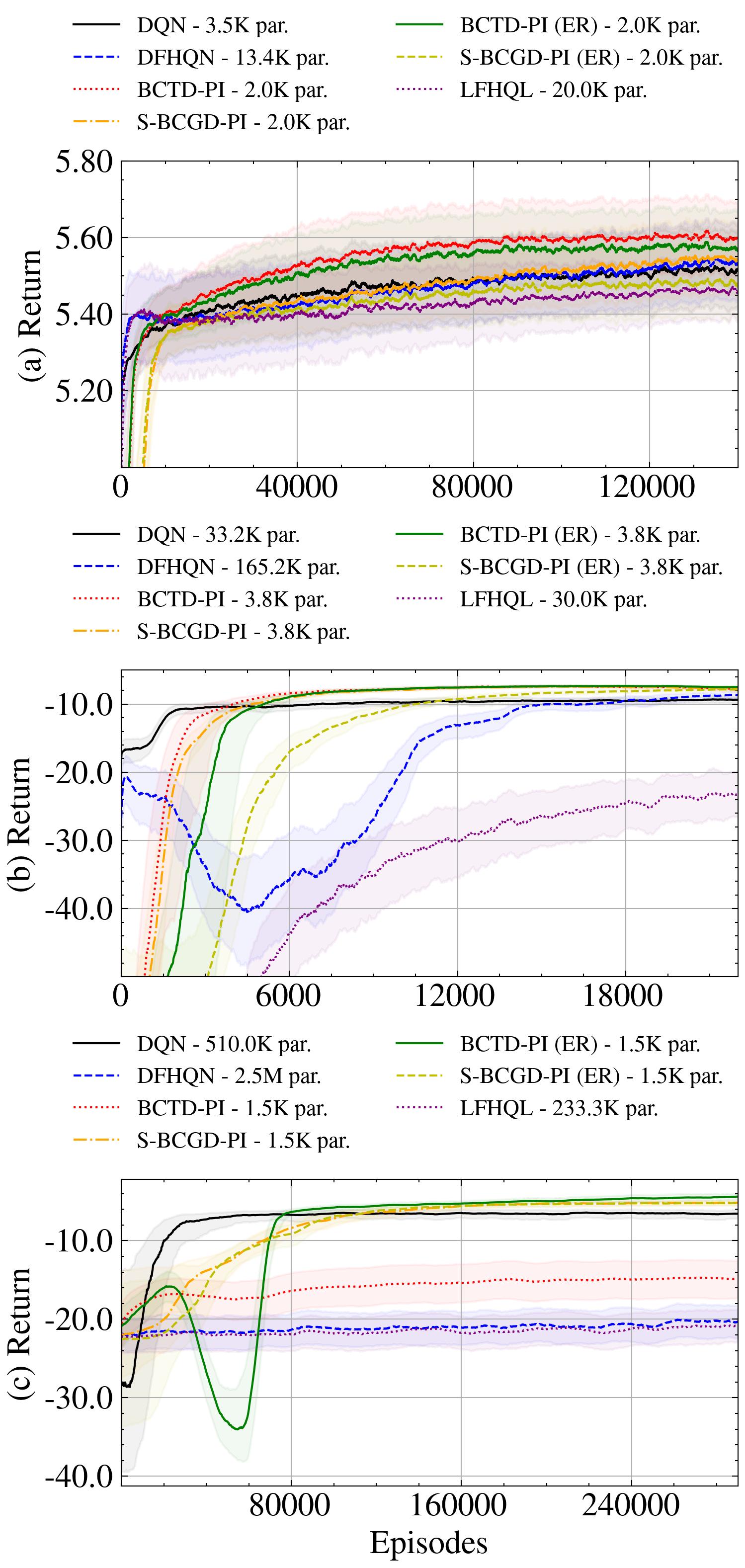}
    \vspace{-.4cm}
    \caption{Average return for \textbf{S-BCGD-PI} and \textbf{BCTD-PI} against different baselines for (a) the wireless comms., (b) the battery-charging, (c) and the source-coding setups. \textbf{S-BCGD-PI} and \textbf{BCTD-PI} converge faster while requiring less params.}
    \label{fig::resource_res}
    \end{center}
    \vspace{-.6cm}
\end{figure}

\noindent \textbf{Convergence Properties.} 
{
We evaluate both PE and PI. In the PE setting, the goal is to recover the VFs of the optimal policy. The corresponding optimal solutions are obtained by solving the BEqs via policy iteration, which we use as ground truth. The entries of the initial factors for \textbf{BCD-PE} and \textbf{BCGD-PE} are sampled from a uniform distribution.
Fig.~\ref{fig::grid_res}-a and Fig.~\ref{fig::grid_res}-b show the PE results for different ranks. \textbf{BCD-PE} achieves zero error, both in terms of NFE relative to the true VFs and in terms of $L(\ccalQ)$, indicating exact recovery of the optimal VFs. 
In contrast, \textbf{BCGD-PE} does not reach zero error, suggesting convergence to local optima. Although both methods show monotonically decreasing errors, consistent with Theorems \ref{thm::bcd_convergence} and \ref{thm::bcgd_convergence}, only \textbf{BCD-PE} recovers the true solution.
This motivates evaluating the full PI pipeline to determine whether such approximation errors compound or are corrected during policy improvement.
We fix the PE step to $M=10$ iterations and analyze \textbf{BCD-PI} and \textbf{BCGD-PI}. 
As shown in Fig.~\ref{fig::grid_res}-c, \textbf{BCD-PI} achieves zero NFE, meaning that it recovers the optimal VFs and, consequently, the optimal policies. 
In contrast, \textbf{BCGD-PI} does not attain zero NFE, reflecting the residual approximation error observed in the PE stage.
However, the ultimate objective is not perfect VF estimation but strong policy performance. Fig.~\ref{fig::grid_res}-d shows the empirical return of the induced policies. Despite its imperfect VF estimation, \textbf{BCGD-PI} produces policies whose performance remains close to optimal, hovering around the optimal return (yellow line), while requiring less computational effort per iteration. \textbf{BCD-PI}, on the other hand, consistently matches the optimal return.
Importantly, even when PE is not exact, strong policies can still be recovered. For example, \textbf{BCD-PI} achieves optimal performance with a rank-15 policy using less than half the parameters, despite attaining only $28\%$ NFE. 
%This confirms that exact VF recovery is not a prerequisite for optimal control, and moderate approximation error in the VF may still induce an optimal policy.
}

\subsection{Stochastic policy iteration in Gymnasium}
\label{SS:res_gym}

{
We evaluate stochastic policy iteration, as outlined in Algorithm \ref{alg::policy_iteration}, using \textbf{S-BCGD-PI} and \textbf{BCTD-PI} in classical control problems where the MDP model is not available. Specifically, we consider the Pendulum and CartPole environments from the Gymnasium benchmark \cite{towers2024gymnasium}.
We compare the proposed algorithms against standard baselines: finite-horizon TD-learning with linear function approximation (\textbf{LFHQL}) \cite{geramifard2013tutorial}, where radial basis functions are used as features \cite{keller2006automatic}; deep $Q$-learning (\textbf{DQN}) \cite{mnih2015human}, implemented with a multi-layer perceptron (MLP); and its finite-horizon counterpart \textbf{DFHQN} \cite{de2020fixed}, also implemented with an MLP. As customary in \textbf{DQN} and \textbf{DFHQN}, we employ ER buffers with capacity $10$ for Pendulum and $100$ for CartPole. 
To optimize the baseline methods, we used the Adam optimizer \cite{kingma2014adam}. Its hyperparameters were tuned for each method to achieve the fastest convergence while ensuring strong performance in terms of returns.
More architectural details are provided in the repository \cite{rozada2024code}.
Regarding the environments, Pendulum and CartPole are not finite-horizon tasks, as their reward does not explicitly depend on time and the control objective is stationary. Since our focus is on time-critical decision making, we modify them to introduce explicit horizon dependence. We set a short horizon $H=10$ to emphasize the critical role of time. Both systems are initialized in the upright position and the reward is provided only at the terminal time step. Consequently, early actions have a strong impact on the outcome, forcing the agent to account for the remaining horizon. %This makes time an essential component of the problem and aligns the environments with the finite-horizon setting.
}

{
Performance is evaluated in terms of episodic return. Hyperparameters are tuned via exhaustive search with the goal of minimizing model complexity across all methods. For our algorithms, the rank $K$ is treated as an additional hyperparameter and tuned accordingly. Each method is tested over $100$ independent runs, with $4{,}000$ episodes per run for both environments. Exploration is handled via an $\varepsilon$-greedy strategy. For \textbf{S-BCGD-PI} and \textbf{BCTD-PI}, we set $M=1$, updating the policy after every transition. To ensure fair comparison, we match the ER buffer capacities used by the NN-based baselines and additionally evaluate our methods without ER.
The median returns are shown in Fig.~\ref{fig::gym_res}-a and Fig.~\ref{fig::gym_res}-b. Overall, the proposed methods require fewer parameters than the neural baselines while achieving faster empirical convergence. In the Pendulum experiment, \textbf{DQN} initially converges quickly but attains inferior performance, as it does not explicitly account for the time dimension. \textbf{DFHQN} eventually surpasses \textbf{DQN} and reaches performance comparable to our methods, although it requires more episodes to do so. Among the proposed algorithms, \textbf{BCTD-PI}, both with and without ER, exhibits the fastest convergence across all methods. \textbf{S-BCGD-PI} also converges faster than \textbf{DFHQN}, confirming the effectiveness of the proposed stochastic low-rank PI framework.
}

\subsection{Stochastic policy iteration in realistic setups}
\label{SS:res_stochastic}

{
We further evaluate the performance of \textbf{S-BCGD-PI} and \textbf{BCTD-PI} in more realistic scenarios, again in the absence of an available MDP model. Specifically, we consider a wireless communication problem, a battery charging problem, and a source coding problem. These settings are characterized by limited time horizons, where managing time effectively is critical to performance.
As in the previous subsection, we compare the proposed algorithms against \textbf{LFHQL}, \textbf{DQN}, and \textbf{DFHQN}. %We begin by describing the problems.
}

\vspace{0.1cm}
\noindent \textbf{Wireless setup.} This setup  considers a time-constrained opportunistic multiple-access wireless system. A single user, equipped with a battery and a queue, transmits packets to an access point over $C = 2$ orthogonal channels within a finite horizon of $H = 4$ time steps. The user accesses the channels opportunistically, as each channel may be occupied.
The system state comprises: (a) the fading level and occupancy state of the $C = 2$ channels, (b) the battery energy level, and (c) the number of packets in the queue. At each time step, the user observes the state and selects the discretized transmission power for each channel. This leads to a $9$-dimensional tensor containing $16$ million entries. The transmission rate is determined by Shannon's capacity formula, and a packet loss of $80\%$ occurs if the channel is occupied.
The reward function combines a weighted sum of the battery level with a positive weight, and the remaining packets in the queue with a negative weight. This structure creates a trade-off between maximizing throughput and conserving battery energy. Early in the episode, the agent can be conservative, waiting for high-SNR, unoccupied channels. However, as the time horizon shrinks, clearing the queue takes precedence, leading to less efficient transmissions that deplete the battery.

\noindent \textbf{Battery charging setup.} A single battery is connected to multiple energy sources, including solar, wind, and grid power, with varying costs and availability. The time horizon is limited to $H=5$ time-steps, requiring decisions about charging rates and energy source selection to balance energy demands, costs, and battery durability.
The state is defined by: (a) the state of charge (SoC) of the battery; (b) the availability of renewable energy sources; and (c) the grid energy price, which fluctuates over time. At each time step, the agent selects charging rates for each energy source. Solar and wind charging rates depend on the availability, while grid charging incurs a cost. This leads to a $7$-dimensional tensor with $10$ million entries.
The reward penalizes charging costs and battery degradation, and includes an additional penalty for failing to reach the target SoC. This creates a tension between immediate and long-term objectives. In earlier time steps, the agent should prioritize renewable sources to minimize costs and degradation. However, as the horizon approaches its end, meeting the target SoC becomes critical, even at the expense of higher costs or durability penalties.

\noindent \textbf{Source-channel coding setup.} {
This setup considers a finite-horizon source–channel coding problem with $C = 4$ stochastic channels over a horizon of $H = 5$ time steps. 
The agent transmits an initial amount of data while dynamically selecting a compression level and a normalized power allocation across channels.
The state comprises the remaining data and the instantaneous channel gains, which evolve according to a Gauss–Markov process. 
The action consists of selecting a compression level and power allocation across channels. Compression reduces the remaining data at the cost of distortion, while the allocation determines the achievable rate following a $\log(1+\mathrm{SNR})$ model with stochastic transmission success.
Including time as an additional dimension, the finite-horizon VF forms an $11$-dimensional tensor with above $1$ billion entries, highlighting the scale of the problem.
The reward penalizes distortion and transmission cost at each step and applies a terminal penalty proportional to the remaining data. This creates a  time-dependent trade-off: early decisions may favor conservative transmission, whereas near the end of the horizon clearing the remaining data becomes critical, even at higher cost.
}

{
We again evaluate all methods in terms of episodic return, with hyperparameters tuned via exhaustive search to minimize model complexity. 
Each algorithm is tested over $100$ independent runs. In the wireless problem, each run consists of $140{,}000$ episodes; in the battery charging problem, $20{,}000$ episodes; and in the source coding problem, $30{,}000$ episodes. For both \textbf{S-BCGD-PI} and \textbf{BCTD-PI}, we set $M=1$, updating the policy after every transition. We additionally evaluate variants with and without an ER buffer (capacity $4$ in the wireless setup, $5$ in battery charging, and $100$ in source coding). Median returns are reported in Fig.~\ref{fig::resource_res}.
Across all three problems, the proposed tensor-based methods require significantly fewer parameters than the neural baselines and show faster empirical convergence. \textbf{DQN} typically converges quickly but attains inferior performance, as it ignores the time dimension. \textbf{DFHQN}, which explicitly accounts for time, eventually reaches performance comparable to our methods, although it requires more episodes and samples to do so. Among the proposed algorithms, \textbf{BCTD-PI} consistently achieves the fastest convergence. \textbf{S-BCGD-PI} performs competitively, though its behavior depends on the use of ER. In some settings ER slightly slows convergence or marginally reduces the final return, while in others it stabilizes training.
Regarding the number of required parameters, the contrast between the proposed methods and the baselines becomes particularly pronounced in the source coding problem. 
The action space grows exponentially with the number of action dimensions, and in that setup we have $5$ action dimensions. 
As a result, for the MLP-based approaches the size of the readout layer becomes large. In the case of \textbf{DFHQN}, this leads to more than $2$ million parameters. In contrast, the tensor parameterization keeps the number of factor entries controlled, allowing scalable learning despite the combinatorial growth of the action space. 
Notably, in this more demanding setting, \textbf{S-BCGD-PI} without ER fails to converge.
}

%%%%%%%%%%%%%%%%%%%%%%%%%%%%%%%%%%%%%%%%%%

%%%%%%%%%%%%%%%%%%%%%%%%%%%%%%%%%%%%%%%%%%
\section{Conclusion}
\label{S:conclusion}

This paper presents an approach to estimate the optimal VFs of finite-horizon MDPs by modeling them as a low-rank multi-dimensional tensor, with time as one of its dimensions. Using the PARAFAC decomposition, our method ensures that the model size grows linearly with the tensor dimensions.
To estimate the optimal VFs, we introduced a low-rank optimization framework based on the BEqs and policy iteration. For MDPs with known models, we proposed two BC algorithms, BCD-PE and BCGD-PE, both of which converge to stationary points. For unknown MDP models, we developed two stochastic methods: S-BCGD-PE, which guarantees convergence to a stationary point, and BCTD-PE, which leverages unbiased estimates of the optimization problem.
We evaluated our methods on a grid-world setup with a known MDP model and two resource allocation problems: a wireless communication setup and a battery charging task. Our algorithms are significantly more efficient in sample usage and parameter count.
Future work will explore integrating tensor low-rank methods with NNs, as we believe these properties hold great potential for designing architectures for dynamical systems.

%This paper introduced an approach to estimate the optimal VFs of finite-horizon MDPs by modeling VFs as a low-rank multi-dimensional tensor, with time being one of the dimensions of the tensor. Leveraging the PARAFAC decomposition, our approach ensures that the model size grows linearly with the tensor dimensions, keeping the number of parameters manageable.
%In order to estimate the optimal VFs we presented a low-rank optimization framework rooted in the BEqs and policy iteration.  To address the optimization for MDPs with known models, we proposed two BC algorithms, BCD-PE and BCGD-PE. Both algorithms converge to stationary points. For unknown MDP models, we developed two stochastic methods: S-BCGD-PE, which guarantees convergence to a stationary point, and BCTD-PE, which uses unbiased estimates for the underlying optimization problem.
%The proposed methods were tested on a grid-world setup with a known MDP model and two resource allocation problems: a wireless communication setup and a battery charging task. Compared to linear and NN methods, our algorithms achieved comparable returns while being significantly more efficient in sample usage and parameter count.
%Future work will explore integrating tensor low-rank methods with NNs, as we believe low-rank properties hold great potential for designing architectures for dynamical systems.

%%%%%%%%%%%%%%%%%%%%%%%%%%%%%%%%%%%%%%%%%%

%%%%%%%%%%%%%%%%%%%%%%%%%%%%%%%%%%%%%%%%%%
\appendices

\section{Proof of Lemma~\ref{prop::least_squares}}
\label{S:prop_least_squares}

{Recall that the Bellman-error objective is
\(
L(\ccalQ)
=
\frac{1}{H}
\sum_{h=1}^{H}
\sum_{s,a}
\delta_\ccalQ(s,a,h)^2.
\)

\paragraph{Step 1: Vectorized Bellman residuals}

For each time step $h$, define the vectorized value slice
\[
\bar{\bbq}_h = \vvec(\tenbQ(:,:,h)) \in \mathbb{R}^{|\mathcal{S}||\mathcal{A}|},
\]
and let $\bbr \in \mathbb{R}^{|\mathcal{S}||\mathcal{A}|}$ denote the reward vector.
Then the Bellman residual at time $h$ can be written as
\(
\bbr + \bbP^{\pi_h} \bar{\bbq}_{h+1} - \bar{\bbq}_h,
\)
so that
\begin{equation}
L(\ccalQ)
=
\frac{1}{H}
\sum_{h=1}^{H}
\|
\bbr + \bbP^{\pi_h} \bar{\bbq}_{h+1} - \bar{\bbq}_h
\|_2^2.
\label{eq::appendix_L_vector_form}
\end{equation}

\paragraph{Step 2: Linear dependence on a single factor}

Fix a mode $d$ and keep all other factors fixed.
We now show that each $\bar{\bbq}_h$ depends linearly on $\bbq_d=\vvec(\bbQ_d^\Tr)$.

\medskip
\noindent
\textbf{Case 1: $d < D$.} Using properties of the PARAFAC decomposition and vectorization,
\[
\bar{\bbq}_h
=
\Big(\bigodot_{i=1}^{D-1} \bbQ_i\Big)
\bbQ_D(h,:)^\Tr.
\]

Rearranging terms to isolate $\bbQ_d$ and applying standard Kronecker and Khatri–Rao identities yields
\[
\bar{\bbq}_h
=
\bbC^h_{d^{\setminus}} \bbq_d,
\]
where $\bbC^h_{d^{\setminus}}$ is defined as
\[
\bbC^h_{d^{\setminus}}
=
\bbT_d
\Big(
\bbI_d \otimes
\Big(
\Big(\bigodot_{i\neq d} \bbQ_i\Big)
\odot \bbQ_D(h,:)
\Big)
\Big),
\]
which matches the definition provided in the main text.

\medskip
\noindent
\textbf{Case 2: $d = D$.} In this case,
\[
\bar{\bbq}_h
=
\Big(\bigodot_{i=1}^{D-1} \bbQ_i\Big)
\bbQ_D(h,:)^\Tr.
\]
Writing $\bbQ_D(h,:)^\Tr = \bbQ_D^\Tr \bbe_h$,
we obtain
\[
\bar{\bbq}_h
=
\bbT_D
\Big(
\bbe_h^\Tr \otimes
\Big(\bigodot_{i=1}^{D-1} \bbQ_i\Big)
\Big)
\bbq_D
=
\bbC^h_{D^{\setminus}} \bbq_D,
\]
which coincides with the second branch of the definition of $\bbC^h_{d^{\setminus}}$.

\paragraph{Step 3: Bellman residual as a linear function of $\bbq_d$}

Substituting $\bar{\bbq}_h = \bbC^h_{d^{\setminus}} \bbq_d$ into
\eqref{eq::appendix_L_vector_form} gives
\[
L(\ccalQ)
=
\frac{1}{H}
\sum_{h=1}^{H}
\|
\bbr
+
\bbP^{\pi_h}
\bbC^{h+1}_{d^{\setminus}} \bbq_d
-
\bbC^h_{d^{\setminus}} \bbq_d
\|_2^2.
\]

Define
\[
\bar{\bbC}^h_{d^{\setminus}}
=
\bbP^{\pi_h}
\bbC^{h+1}_{d^{\setminus}}
-
\bbC^h_{d^{\setminus}}.
\]

Then
\[
L(\ccalQ)
=
\frac{1}{H}
\sum_{h=1}^{H}
\|
\bbr
+
\bar{\bbC}^h_{d^{\setminus}} \bbq_d
\|_2^2.
\]

\paragraph{Step 4: Stacked LS form}

Let
\(
\bar{\bbr}
=
[\bbr^\Tr,\dots,\bbr^\Tr]^\Tr
\in
\mathbb{R}^{|\mathcal{S}||\mathcal{A}|H},
\)
and define the block-stacked matrix:
\begin{equation}
\label{eq::c_def}
\bar{\bbC}_{d^{\setminus}}
=
\left[
(\bbP^{\pi_1}\bbC^2_{d^{\setminus}} - \bbC^1_{d^{\setminus}})^\Tr,
...,
(\bbP^{\pi_H}\bbC^{H+1}_{d^{\setminus}} - \bbC^H_{d^{\setminus}})^\Tr
\right]^\Tr\!\!.
\end{equation}

Stacking the $H$ residual vectors yields
\[
L(\ccalQ)
=
\frac{1}{H}
\|
\bar{\bbr}
+
\bar{\bbC}_{d^{\setminus}}
\bbq_d
\|_2^2.
\]

This shows that, when all other factors are fixed, minimizing $L(\ccalQ)$ w.r.t. $\bbQ_d$ reduces to a standard LS problem in $\bbq_d$.}

\hfill $\square$

\section{Proof of Theorem~\ref{thm::bcd_convergence}}
\label{S:app_bcd_conv}

This proof will show that the convergence of BCD-PE follows from \cite[Th. 2.8]{xu2013block}, which addresses problems of the form
\begin{align}
    \label{eq::problem_ref_thm}
    \min_{\bar \bbx \in \ccalX} F(\bar \bbx) = f_0(\bbx_1, \hdots, \bbx_D) + \sum_{d=1}^D f_d(\bbx_d).
\end{align}
The variable $\bar \bbx$ is decomposed into $D$ blocks $\bbx_1, \hdots, \bbx_D$. The function $f_0$ is block multi-convex, though potentially non-convex, and each $f_d$ is a convex function. Clearly, \eqref{eq::problem} is a specific instance of \eqref{eq::problem_ref_thm}. Specifically, the optimization variables are the factors $\ccalQ$, whose vectorizations are stacked as $\bar \bbx = [\bbq_1^\Tr , \hdots , \bbq_D^\Tr]^\Tr  \in \reals^{\left(\sum_{d=1}^D |\ccalD_d|\right)K^D}$, with $\bbq_d \in \reals^{|\ccalD_d|K}$. The functions are  $f_d(\bbq_d) = 0$, rendering $F(\bar \bbx) = f_0(\bbx_1, \hdots, \bbx_D) = L(\ccalQ)$.
The update rule of BCD-PE defined in \eqref{eq::q_factor_iterates_rule} is an instance of the update in \cite[Eq. (1.3a)]{xu2013block}, which is proven to converge provided that 6 assumptions are verified \cite[Th. 2.8]{xu2013block}. Next, we list the assumptions and show that our problem satisfies them.

\begin{assumption}
    \label{ass::continuity}
    $L$ is continuous in $\text{dom}(L)$ and satisfies $\text{inf}_{\ccalQ \in \text{dom}(L)}L(\ccalQ) > - \infty$.
\end{assumption}

\begin{assumption}
    \label{ass::nash_point}
    The set of stationary points of $L$ is non-empty.
\end{assumption}

\begin{assumption}
    \label{ass::KL_inequality}
    $L$ satisfies the Kurdyka–Łojasiewicz (KL) inequality at the stationary points.
\end{assumption}

\begin{assumption}
    \label{ass::strong_convexity}
    There exist constants $0 < \ell^0_d  < \infty$, for $d=1, \hdots, D$, such that 
    \begin{equation}
        \notag
        L(\{ \bbQ^{(m)}_1, \hdots, \bbQ^{(m)}_{d-1}, \bbQ_d, \bbQ^{(m - 1)}_{d+1}, \hdots, \bbQ^{(m-1)}_D \})
    \end{equation}
    is $\ell^0_d$-strongly convex w.r.t. $\bbQ_d$.
\end{assumption}

\begin{assumption}
    \label{ass::lipschitz_cont_gradient}
    The gradient $\nabla_{\bbQ_d} L(\ccalQ)$ is $\ell_d^1$-Lipschitz-continuous on any bounded set with $\ell_d^1 \geq \ell_d^0$.
\end{assumption}

\noindent \textit{Ass. \ref{ass::continuity}} is satisfied, since $L(\ccalQ)$ is an addition of norms. 

\vspace{.1cm}
\noindent \textit{Ass. \ref{ass::nash_point}} is verified by showing that $L(\ccalQ)$ is differentiable and coercive. The former holds, and the latter does too because, from \eqref{eq::problem_fixed_least_squares}, it follows that if $\| \bbQ_d \|_F^2 \to \infty$, then $L(\ccalQ) \to \infty$.

\vspace{.1cm}
\noindent \textit{Ass. \ref{ass::KL_inequality}} is also satisfied, as $L(\ccalQ)$ is a multi-linear polynomial function and, therefore, real analytic, which is a class of functions  known to satisfy the KL property.

\vspace{.1cm}
\noindent We move now to show that \textit{Ass. \ref{ass::strong_convexity} and \ref{ass::lipschitz_cont_gradient}} are satisfied, which requires a longer proof. 
Let $\bbA \preceq \bbB$ indicate that $\bbB - \bbA$ is positive definite. To verify Ass. \ref{ass::strong_convexity}, it suffices to show that for all $d$, it holds $\ell_d^0 \bbI \preceq \nabla^2_{\bbq_d} L(\ccalQ)$, while for Ass. \ref{ass::lipschitz_cont_gradient} one must show that $\nabla^2_{\bbq_d} L(\ccalQ) \preceq \ell_d^1 \bbI$ holds for all $d$. Since in this particular case we have $\nabla^2_{\bbq_d} L(\ccalQ) = \bar{\bbC}_{d^{\setminus}}^\Tr \bar{\bbC}_{d^{\setminus}}$, to prove Ass. \ref{ass::strong_convexity} and \ref{ass::lipschitz_cont_gradient}, we have to show that
\begin{align}\label{eq::HessianUpperAndLowerBoundedAssumptions8_and_9}
       \ell_d^0 \bbI \preceq \bar{\bbC}_{d^{\setminus}}^\Tr \bar{\bbC}_{d^{\setminus}} \preceq \ell_d^1 \bbI.
\end{align}
To ensure $\ell_d^0 \bbI \preceq \bar{\bbC}_{d^{\setminus}}^\Tr \bar{\bbC}_{d^{\setminus}}$, it suffices to show that the smallest eigenvalue of $\bar{\bbC}_{d^{\setminus}}^\Tr \bar{\bbC}_{d^{\setminus}}$ is nonzero, which holds if $\bar{\bbC}_{d^{\setminus}}$ is full-column rank for all $d$.
%We show this separately for (a) $d < D$ and (b) $d = D$.  

\vspace{.1cm}
\noindent Let us focus first on \textit{the case (a) where} $d < D$. The tall matrix $\bar{\bbC}_{d^{\setminus}}$ is composed of the block matrices $\bar \bbC^h_{d^{\setminus}}$ stacked column-wise. Since the number of columns of $\bar{\bbC}_{d^{\setminus}}$ and $\bar \bbC^h_{d^{\setminus}}$ is $|\ccalD_d|K$ for both matrices, showing that one of these blocks is full-column rank is sufficient to prove that $\bar{\bbC}_{d^{\setminus}}$ is full-column rank. We can then consider the block $\bar \bbC^H_{d^{\setminus}}$, which is defined as
\begin{align}
    \notag
    \bar \bbC^H_{d^{\setminus}} = \bbP^{\pi_h} \bbC^{H+1}_{d^{\setminus}} - \bbC^{H}_{d^{\setminus}} = - \bbC^{H}_{d^{\setminus}},
\end{align}
since, by convention, the $Q$-values for $H+1$, and therefore, $\bbQ_D(H+1, :) = \bbzero$.
The key task now is to prove that
$$\bbC^{H}_{d^{\setminus}}= \bbT_d (\bbI_d \otimes (({\textstyle \bigodot_{i=1 \neq d}^{D-1}} \bbQ_i ) \odot \bbQ_D(H, :)))$$
is full-column rank. This holds because: (i) $\bbT_d$ is a permutation matrix that merely reorders rows, thereby preserving the rank of the matrix; and (ii) the Khatri-Rao product of the factors is full-column rank. The latter follows from two key points: first, the factors themselves are full-column rank, as stated in Ass. \ref{ass::regularity}; and second, the Khatri-Rao product of two full-column rank matrices is almost surely full-column rank \cite[Cor. 1]{jiang2001almost}.

\vspace{.1cm}
\noindent 
Having established that $\bar{\bbC}_{d^{\setminus}}$ is full-column rank for all $d < D$, we now proceed to show that the same property holds for \textit{the case (b) where} $d = D$, i.e., $\bar{\bbC}_{D^{\setminus}}$ is also full-column rank.
In this case, the argument differs because the blocks $\bbC^h_{D^{\setminus}}$ are not full rank for all $h$. To prove that $\bar{\bbC}_{D^{\setminus}}$ is full-column rank, we use a recursive argument. For each block of $\bar{\bbC}_{D^{\setminus}}$ we have
\begin{equation}
    \notag
    \bar{\bbC}^h_{D^{\setminus}} = \bbP^{\pi_h} \bbC^{h+1}_{D^{\setminus}} - \bbC^{h}_{D^{\setminus}}.
\end{equation}
Recall that for $d = D$, the matrix $\bbC^{h}_{D^{\setminus}} \in \reals^{(\prod_{d=1}^{D-1} |\ccalD_d|) \times HK}$ is defined in \eqref{eq::c_def} as $\bbT_D (\bbe_h^\Tr \otimes (\bigodot_{d=1}^{D-1} \bbQ_d))$. The presence of the canonical vector $\bbe_h$ implies that only $K$ columns are non-zero. These columns are grouped into $H$ blocks of $K$ columns each, where only the block corresponding to $h$ contains non-zero entries. Specifically, this block contains $\bigodot_{d=1}^{D-1} \bbQ_d$.
Consider the case where $h = H$. By Ass. \ref{ass::regularity}, \cite[Cor. 1]{jiang2001almost} and the fact of factors having $K$ columns, $\bbC^{H}_{D^{\setminus}}$ has $K$ independent columns. Consequently, $\bar{\bbC}^{H}_{D^{\setminus}}$ also has $K$ independent columns. Next, consider $\bar{\bbC}^{H-1}_{D^{\setminus}}$, which shows the same block structure. It has two non-zero blocks located at positions $H-1$ and $H$. 
By a similar reasoning, $\bar{\bbC}^{H-1}_{D^{\setminus}}$ also has at least $K$ independent columns. Furthermore, the stacking
\begin{equation}
    \notag
    \begin{bmatrix}
        (\bar \bbC^{H-1}_{D^{\setminus}})^\Tr, 
        (\bar \bbC^{H}_{D^{\setminus}})^\Tr
    \end{bmatrix}^\Tr \in \reals^{2 \left( \prod_{d=1}^{D-1} |\ccalD_d| \right) \times HK} 
\end{equation}
has rank $2K$. Extending this argument to include the block $\bar \bbC^{H-2}_{D^{\setminus}}$, which has non-zero blocks in $H-2$ and $H-1$, the rank increases to $3K$. 
Repeating this, we conclude that $\bar{\bbC}_{D^{\setminus}}$ has rank $HK$, which is the number of columns of $\bar{\bbC}_{D^{\setminus}}$.
Therefore, since $\bar{\bbC}_{d^{\setminus}}$ is full-column rank for both $d<D$ and $d=D$, the left hand side inequality in \eqref{eq::HessianUpperAndLowerBoundedAssumptions8_and_9} $\ell_d^0 \bbI \preceq \bar{\bbC}_{d^{\setminus}}^\Tr \bar{\bbC}_{d^{\setminus}}$ holds for all $d$, with $\ell_d^0$ being the smallest eigenvalue of $\bar{\bbC}_{d^{\setminus}}^\Tr \bar{\bbC}_{d^{\setminus}}$.

Finally, we aim to establish that $\bar{\bbC}_{d^{\setminus}}^\Tr \bar{\bbC}_{d^{\setminus}} \! \preceq \! \ell_d^1 \bbI$, the right hand side inequality of \eqref{eq::HessianUpperAndLowerBoundedAssumptions8_and_9}, holds for all $d$. 
A key step is to show that the sequence $\{\ccalQ^{(m)}\}_{m \geq 1}$ generated by BCD-PE is bounded. This follows from the descent properties of BCD-PE and the normalization after each iteration.
Consider bounded initial factors $\ccalQ^{(0)}$. In the first iteration, the gradients $\nabla_{\bbq_d} L(\ccalQ^{(1)}_d) = (\bbC^{(1)}_{d^\setminus})^\Tr (\bar \bbr + \bbC^{(1)}_{d^\setminus} \bbq_d)$ remain bounded for all $d$. Thus, BCD-PE satisfies all assumptions locally, ensuring a decrease in the loss, i.e., $L(\ccalQ^{(1)}) \! < \! L(\ccalQ^{(0)})$ (see Lemma 2.2 in \cite{xu2013block}). Since $L(\mathcal{Q})$ is coercive and continuous, the descent property guarantees that the tensor iterates $\tenbQ^{(m)}=[[\mathcal{Q}^{(m)}]]$ remain bounded. However, due to the scaling-counter-scaling ambiguity in PARAFAC, a bounded tensor iterate does not directly imply bounded factor matrices, as one factor may grow unbounded while another shrinks proportionally.
Nonetheless, since our algorithm includes a normalization step ensuring $\|\bbQ_1^{(m)}\|_F \! = \! ... \! = \! \|\bbQ_D^{(m)}\|_F$, the boundedness of $\tenbQ^{(m)} \! = \! [[\mathcal{Q}^{(m)}]]$ implies that the factor matrix iterates are also bounded.
Consequently, there exists a constant for all $d$ and $m$ such that $(\bar{\bbC}_{d^{\setminus}}^{(m)})^\Tr \bar{\bbC}_{d^{\setminus}}^{(m)} \preceq \ell_d^1 \bbI$, which proves Ass. \ref{ass::strong_convexity} and \ref{ass::lipschitz_cont_gradient}.

\section{Proof of Theorem~\ref{thm::bcgd_convergence}}
\label{S:app_bcgd_conv}
The structure of the proof is similar to that in Appendix \ref{S:app_bcd_conv}. Here, the BCGD-PE update rule in \eqref{eq::bcgd_rule} corresponds to the prox-linear update in \cite[Eq. (1.3c)]{xu2013block}, with extrapolation weights $\omega^{(m)}_d = 0$ for all $m$ and $d$. According to \cite[Th. 2.8]{xu2013block}, convergence is guaranteed under Ass. \ref{ass::continuity}, \ref{ass::nash_point}, \ref{ass::KL_inequality}, and \ref{ass::lipschitz_cont_gradient}. Since Appendix \ref{S:app_bcd_conv} establishes that these assumptions hold for our problem, the convergence of BCGD-PE is ensured.

{
{

\section{Proof of Theorem \ref{thm::perf_bound}}
\label{SS:performance_bound}
Let $\widehat{Q}_h^{(n)}(s,a):=\tenbQ^{(n)}(\bbi_s,\bbi_a, h)$ denote the tensor approximation of $Q_h^{\pi^{(n)}}(s,a)$ from the evaluation step using BCD-PE or BCGD-PE.
We begin by using the performance difference lemma for finite-horizon MDPs \cite[Lemma 3.1]{agnihotri2024cop}:
\begin{align}
\label{eq::perf_diff_lemma}
&J(\pi^{(n+1)}) - J(\pi^{(n)})
= \\
\notag
&\sum_{h=1}^H
\mathbb{E}_{s_h \sim P_h^{\pi^{(n+1)}}}
\left[
Q_h^{\pi^{(n)}}(s_h,\pi_h^{(n+1)}(s_h))
-
V_h^{\pi^{(n)}}(s_h)
\right].
\end{align}

\begin{align}
    g_h^{(n)}(s) = Q_h^{\pi^{(n)}}(s, a_h^\star(s)) - Q_h^{\pi^{(n)}}(s, \pi_h^{(n)}(s)),
\end{align}
Now, recall the definition of the greedy advantage and consider the definition of the approximation error
\begin{align}
    \notag
    &g_h^{(n)}(s) = Q_h^{\pi^{(n)}}(s, a_h^\star(s)) - Q_h^{\pi^{(n)}}(s, \pi_h^{(n)}(s)),
    \\
    \notag
    &\varepsilon_h^{(n)}(s) = Q_h^{\pi^{(n)}}(s, a_h^\star(s)) - Q_h^{\pi^{(n)}}(s, \pi_h^{(n+1)}(s)).
\end{align}
Then
\(
Q_h^{\pi^{(n)}}(s,\pi_h^{(n+1)}(s))
-
V_h^{\pi^{(n)}}(s)
=
g_h^{(n)}(s)
-
\varepsilon_h^{(n)}(s),
\)
and substituting into \eqref{eq::perf_diff_lemma} gives
\begin{align}
\label{eq::perf_decomp_eps}
J(\pi^{(n+1)}) - J(\pi^{(n)})
=
\sum_{h=1}^H
\mathbb{E}_{s_h}
\left[
g_h^{(n)}(s_h)
-
\varepsilon_h^{(n)}(s_h)
\right].
\end{align}

Now, we define the $h$-dependent greedy action $\hat a_h(s):=\argmax_a \widehat{Q}_h^{(n)}(s,a)$.
Assuming that  $|\widehat{Q}_h^{(n)}(s,a)-Q_h^{\pi^{(n)}}(s,a)|\le e_h^{(n)}$ is bounded for all $(s,a)$, hence
\begin{align}
    Q_h^{\pi^{(n)}}(s,a_h^\star(s))
    &\le \widehat{Q}_h^{(n)}(s,a_h^\star(s)) + e_h^{(n)}, \label{eq:greedy_gap_1}\\
    \widehat{Q}_h^{(n)}(s,a_h^\star(s))
    &\le \widehat{Q}_h^{(n)}(s,\hat a_h(s)), \label{eq:greedy_gap_2}\\
    \widehat{Q}_h^{(n)}(s,\hat a_h(s))
    &\le Q_h^{\pi^{(n)}}(s,\hat a_h(s)) + e_h^{(n)}. \label{eq:greedy_gap_3}
\end{align}
Combining \eqref{eq:greedy_gap_1}--\eqref{eq:greedy_gap_3} yields
\begin{align}
    \notag
    \varepsilon_h^{(n)}(s_h) = Q_h^{\pi^{(n)}}(s,a_h^\star(s)) - Q_h^{\pi^{(n)}}(s,\hat a_h(s)) \le 2e_h^{(n)}
\end{align}
Plugging into \eqref{eq::perf_decomp_eps} yields 
\begin{align}
\label{eq::perf_decomp_err}
J(\pi^{(n+1)}) - J(\pi^{(n)}) \geq \sum_{h=1}^H
\mathbb{E}_{s_h}
\left[g_h^{(n)}(s_h)\right] - 2 \sum_{h=1}^H e_h^{(n)}.
\end{align}

%Now, we recall that we parametrize the state-action VFs through the rank-$K$ tensor model $\tenbQ=[[\ccalQ]]$ and select $\ccalQ$ by minimizing the squared Bellman residual defined in \eqref{eq::problem}
Now, for a fixed iteration $n$ and time-step $h$ we recall the per-step Bellman residual
\begin{equation}
    \notag
    \ell_h^{(n)}(\ccalQ) = \|\bbr + \bbP^{\pi_h^{(n)}}\bbq_{h+1} - \bbq_h\|.
\end{equation}
For the output of the BCD-PE or the BCGD-PE algorithm, $\ccalQ^{(n)}$, this residual can be decomposed as $\ell_h^{(n)}(\ccalQ^{(n)}) = \inf_{\ccalQ\in\ccalF_K}\ell_h^{(n)}(\ccalQ) + \eta_h^{(n)}$, where $\inf_{\ccalQ\in\ccalF_K}\ell_h^{(n)}(\ccalQ)$ captures the approximation error induced by restricting to rank-$K$ tensors and $\eta_h^{(n)}$ captures the optimization error due to not reaching the best rank-$K$ residual.

Now we let $\{ \bbq_{h} \}_{h=1}^H$ the true vectorized VFs, $\{\tilde \bbq_{h} \}_{h=1}^H$ the rank $K$ truncation of the true VFs, and $\{ \bbq^\star_{h} \}_{h=1}^H$ the vectorized minimizers of $\inf_{\ccalQ\in\ccalF_K}\ell_h^{(n)}(\ccalQ)$. Then if follows that
\begin{align}
    &\inf_{\ccalQ\in\ccalF_K}\ell_h^{(n)}(\ccalQ) = \|\bbr + \bbP^{\pi_h^{(n)}}\bbq^\star_{h+1} - \bbq^\star_h\| \\
    \notag
    &\overset{(i)}{\leq} \|\bbr + \bbP^{\pi_h^{(n)}} \tilde \bbq_{h+1} - \tilde\bbq_h\| \\
    \notag
    &\overset{(ii)}= \|\bbr + \bbP^{\pi_h^{(n)}} \tilde \bbq_{h+1} - \tilde\bbq_h - (\bbr + \bbP^{\pi_h^{(n)}} \bbq_{h+1} - \bbq_h) \| \\
    \notag
    &\overset{(iii)}\leq \| \tilde \bbq_{h+1} - \bbq_{h+1}\| + \| \tilde \bbq_{h} - \bbq_{h}\| \leq 2[K^\star - K]_+\beta_{K+1},
\end{align}
where $\beta_{K+1}$ is the infinity norm of the $K+1$-th factor of the true tensor of VFs.
The inequality (i) follows from the suboptimality of $\tilde \bbq$, the equality (ii) from the fact that $\bbq$ satisfy Bellman's equation and thus the terms cancel out, and the last inequality (iii) from Lemma 1 in \cite{rozada2024tensor}. 

Since the only way of having $\ell_h^{(n)}=0$ is $\ccalQ^{(n)}$ being the true VF for time-step $h$, there exists ae constant $\kappa_0 > 0$ such that
\begin{align}
    \notag
    e_h^{(n)} \leq \kappa_0 \ell_{h}^{(n)}(\ccalQ^{(n)}) &= \kappa_0 \left( \inf_{\ccalQ\in\ccalF_K}\ell_h^{(n)}(\ccalQ)+\eta_h^{(n)} \right)\\
    \notag
    &\leq 2 \kappa_0 ((K^\star - K) \beta_{K+1} + \eta_h^{n})
\end{align}
Thus, plugging into \eqref{eq::perf_decomp_err} and absorbing constants into $\kappa_1 > 0$ concludes the proof
\begin{align}
    \notag
    &J(\pi^{(n+1)}) - J(\pi^{(n)}) \geq \\
    \notag
    &\sum_{h=1}^H \mathbb{E}_{s_h \sim P_h^{\pi^{(n+1)}}}\!\left[g_h^{(n)}(s_h)\right]
    -\kappa_1 (K^\star - K) - \kappa_0 \sum_{h=1}^H \eta_h^{(n)},
\end{align}
}

}

\section{Proof of Theorem~\ref{thm::traj_samp_equ}}
\label{S:app_traj_samp_equ}

To show that $L_{\mu^{\pi}}(\ccalQ) = L_{\xi^\pi}(\ccalQ)$ we begin by observing that,
due to the Markov property, the distribution $\mu^\pi$ for trajectories $\tau$ is given by the following expression
\begin{align}
    \notag
    \mu^\pi(\tau) = P_1(s_1) \prod_{i=1}^{H-1} P(s_{i+1} | s_i, \pi_i(s_i)),
\end{align}
where $P_1$ is the probability distribution of the initial step. Next, due to the linearity of expectation we can write
\begin{align}
    \label{eq::expectation_trajectory}
    L_{\mu^\pi}(\ccalQ) \! &= \! \mathbb{E}_{\tau \sim \mu^\pi} \!\! \left[ \frac{1}{H} \!\! \sum_{\sigma_h \in \tau} \tilde \delta_{\ccalQ}(\sigma_h)^2\right] \! = \! \frac{1}{H} \!\! \sum_{\sigma_h \in \tau} \!\! \mathbb{E}_{\tau \sim \mu^\pi} \!\! \left[ \tilde \delta_{\ccalQ}(\sigma_h)^2\right] \!\! .
\end{align}
Then, for each of the $H$ transitions we have
\begin{align}
    \notag
    &\mathbb{E}_{\tau \sim \mu^\pi} \left[ \tilde \delta_{\ccalQ}(\sigma_h)^2\right] = \sum_{\tau \in \ccalT} \tilde \delta_{\ccalQ}(\sigma_h)^2 \mu^\pi(\tau) \\
    \notag
    &=  \sum_{s_1 \in \ccalS} \hdots \sum_{s_H \in \ccalS} \tilde \delta_{\ccalQ}(\sigma_h)^2 \; P_1(s_1) \prod_{i=1}^{H-1} P(s_{i+1} | s_i, \pi_i(s_i))
\end{align}
Now, since $\sigma_h$ depends only on the elements of the $h$-th and $(h+1)$-th time-steps, and the system is Markovian, we  marginalize over the sub-trajectories before and after $h$ and $h+1$ to write
\begin{align}
    \notag
    &\sum_{s_1 \in \ccalS} \hdots \sum_{s_H \in \ccalS} \tilde \delta_{\ccalQ}(\sigma_h)^2 \; P_1(s_1) \prod_{i=1}^{H-1} P(s_{i+1} | s_i, \pi_i(s_i))  \\
    \notag
    &= \!\!\! \sum_{s_h \! \in \ccalS} \! \sum_{\s_{h\!+\!1} \! \in \ccalS} \!\!\! \tilde \delta_{\ccalQ}(\sigma_h)^2 P^\pi_h \! (s_h) P(s_{h+1} \! | s_h, \pi_h(s_h) ) \! = \! \mathbb{E}_{\sigma_h} \!\!  \left[ \tilde \delta_{\ccalQ}(\sigma_h)^2 \right]
\end{align}
Now, by interpreting $1/H$ as $P_\ccalH$, we return to \eqref{eq::expectation_trajectory}  considering $\xi^\pi(\sigma)=P_\ccalH(h) P^\pi_h(s) P(s' | s, \pi_h(s))$  to conclude by observing
\begin{align}
    \notag
    L_{\mu^\pi}(\ccalQ) \! &= \! \mathbb{E}_{\tau \sim \mu^\pi} \!\! \left[ \frac{1}{H} \!\! \sum_{\sigma_h \in \tau} \tilde \delta_{\ccalQ}(\sigma_h)^2\right] \! = \! \frac{1}{H} \!\! \sum_{\sigma_h \in \tau} \!\! \mathbb{E}_{\tau \sim \mu^\pi} \!\! \left[ \tilde \delta_{\ccalQ}(\sigma_h)^2\right] \!\! \\
    \notag
    &= \! \frac{1}{H} \!\! \sum_{\sigma_h \in \tau} \!\! \mathbb{E}_{\sigma_h} \!\! \left[ \tilde \delta_{\ccalQ}(\sigma_h)^2\right] \!\! = \mathbb{E}_{\sigma \sim \xi^\pi} \left[ \tilde \delta_\ccalQ(\sigma)^2\right] = L_{\xi^\pi}(\ccalQ).
\end{align}

\section{Proof of Theorem~\ref{thm::sbcgd_convergence}}
\label{S:app_sbcgd_conv}

The proof shows that S-BCGD-PE satisfies the convergence conditions in \cite[Th. 2.10]{xu2015block}, which addresses the optimization
\begin{equation}
    \label{eq::problem_sto_ref_thm}
    \min_{\bar \bbx \in \ccalX}  \mathbb{E}_{\kappa} \left[ f_0(\bar \bbx, \kappa)\right] + \sum_{d=1}^D f_d(\bbx_d).
\end{equation}
The variable $\bar \bbx$ consists of $D$ disjoint blocks $\bbx_1, \hdots, \bbx_D$, with the domain $\ccalX$ being block multi-convex. The expectation is taken w.r.t. the random variable $\kappa$, and $\mathbb{E}_{\kappa}[f_0(\bar \bbx, \kappa)]$ is continuously differentiable w.r.t. $\bar \bbx$. 
The problem in \eqref{eq::stochastic_problem_samples} is a specific instance of \eqref{eq::problem_sto_ref_thm}, where the optimization variables are the factors $\ccalQ$, vectorized and stacked as $\bar \bbx = [\bbq_1^\Tr; \hdots; \bbq_D^\Tr]^\Tr$. This gives $f_0(\bbx_1, \hdots, \bbx_D) + \sum_{d=1}^D f_d(\bbx_d) = L_{\xi^\pi}(\ccalQ)$, where $\kappa = \sigma$, $\bbx_d=\bbq_d$ for all $d$, and the convex functions $f_d$ are $f_d(\bbx_d) = 0$ for all $d$.
For readability, analogous to the factor iterates $\ccalQ_d^{(m)}$ defined in \eqref{eq::q_factor_iterates}, we define the tensor iterate $\tenbQ^{(m)}_d = [[\ccalQ_d^{(m)}]]$.
Furthermore, we denote the gradient and the stochastic gradient of $L_{\xi^\pi}(\ccalQ)$ as $g_d(\ccalQ) = \nabla_{\bbQ_d} L_{\xi^\pi}(\ccalQ)$ and $\tilde g_d(\ccalQ, \sigma) = \nabla_{\bbQ_d} \tilde \delta_{\ccalQ} (\sigma)^2$, respectively.
The S-BCGD-PE update rule in \eqref{eq::stochastic_bcgd_rule} is a specific case of the update in \cite[Eq. (1.4)]{xu2015block}, which is a stochastic gradient descent step. Since the convergence of \cite[Eq. (1.4)]{xu2015block} is established in \cite[Th. 2.10]{xu2015block}, proving that S-BCGD-PE converges reduces to verifying the same assumptions of \cite[Th. 2.10]{xu2015block}, which are

\begin{assumption}
    \label{ass::sto_lower_bound}
    $L_{\xi^\pi}$ satisfies $\text{inf}_{\ccalQ \in \text{dom}(L_{\xi^\pi})} L_{\xi^\pi}(\ccalQ) > - \infty$.
\end{assumption}

\begin{assumption}
    \label{ass::sto_bounded_gradients}
    The gradient iterates $\mathbb{E}_{\hat \sigma_m} \left[ \| \tilde g_d( \ccalQ_d^{(m)}, \hat \sigma_m) \|^2 \right]$ generated by S-BCGD-PE are upper-bounded for all $d$ and $m$.
\end{assumption}

\begin{assumption}
    \label{ass::sto_lipschitz}
    The gradient $g_d$ is $\ell^0_d$-Lipschitz for some finite $\ell^0_d > 0$ for all $d$.
\end{assumption}

\begin{assumption}
    \label{ass::sto_bias_bound}
    There exists some finite constant $B > 0$ such that for any dimension $d$ and iteration $m$ it holds that $\| \mathbb{E}_{\hat \sigma_m} \left[ \tilde g_d(\ccalQ_d^{(m)}, \hat \sigma_m) - g_d(\ccalQ_d^{(m)})  | \sigma_{m-1}\right] \| \leq B \alpha^{(m)}$.
\end{assumption}

\begin{assumption}
    \label{ass::sto_variance_bound}
    For any dimension $d$ and iteration $m$ it holds that $ \mathbb{E}_{\hat \sigma_m}  \left[ \| \tilde g_d(\ccalQ_d^{(m)}, \hat \sigma_m) - g_d(\ccalQ_d^{(m)}) \| \right] < \infty $.
\end{assumption}

\vspace{.1cm}
\noindent \textit{Ass. \ref{ass::sto_lower_bound}} follows from the fact that $L_{\xi^\pi}$ is the expectation of a quadratic function, which is lower bounded by $0$.

\vspace{.1cm}
\noindent \textit{Ass. \ref{ass::sto_bounded_gradients} and \ref{ass::sto_variance_bound}} hold if $g_d(\ccalQ_d^{(m)})$ and $\tilde{g}_d(\ccalQ_d^{(m)}, \hat{\sigma}_m)$ are upper-bounded for all $d$ and $m$.  
Since $g_d$ and $\tilde{g}_d$, as defined in \eqref{eq::bcgd_rule} and \eqref{eq::stochastic_bcgd_rule}, are polynomial functions of the factors $\ccalQ$, they can grow unbounded if $\ccalQ$ is unbounded.  
However, Ass. \ref{ass::bounded_iterates} ensures that S-BCGD-PE does not cause the factors to grow large.  
Thus, $g_d(\ccalQ_d^{(m)})$ and $\tilde{g}_d(\ccalQ_d^{(m)}, \hat{\sigma}_m)$ remain bounded.

\vspace{.1cm}
\noindent \textit{Ass. \ref{ass::sto_lipschitz}} follows from $g_d(\ccalQ^{(m)})$ being a polynomial function bounded for all $m$. Since $g_d$ is a polynomial function, its gradient $\nabla_{\bbQ_j} g_d(\ccalQ)$ is also a polynomial for all $j=1, ..., D$. Consequently, if $\ccalQ$ is bounded, $\nabla_{\bbQ_j} g_d(\ccalQ)$ is bounded as well for all $j=1, ..., D$. From Ass. \ref{ass::bounded_iterates}, the iterates of S-BCGD-PE are bounded. Therefore, setting $\ell^0_d$ to $\max_{m, j} \|\nabla_{\bbQ_j} g_d(\ccalQ^{(m)})\|$ satisfies Ass. \ref{ass::sto_lipschitz}.

\vspace{.1cm}
\noindent Finally, \textit{Ass. \ref{ass::sto_bias_bound}} requires a longer proof. Simply put, this assumption allows the stochastic gradient to be biased given the previous update, but the bias is of order $\alpha^{(m)}$. As $\alpha^{(m)} \to 0$ with increasing $m$, the bias vanishes.  
%The source of this bias is twofold: first, due to the sampling in the Markovian setup, and second, due to the cyclic nature of the update. Specifically, $\ccalQ_d^{(m)}$ depends on the current transition $\hat \sigma_m$ because the factors $\bbQ^{(m)}_{d'}$ for $d' < d$ have been updated using $\hat \sigma_m$. The use of the ER buffer counteracts the former, while the latter is a consequence of the updates being made on the order of $\alpha^{(m)}$.
To prove this, we begin with some auxilliary derivations. First,  $\tenbQ_d^{(m)}(\bbi_m)$ can be rewritten as
\begin{align}
    %\label{eq::vf_alt_def}
    \label{eq::v_tag}
    v_{d, m} &= \sum_{k=1}^K \prod_{j=1}^{d-1} \bbQ_j^{(m)}(\bbi_m(j), k) \prod_{j=d}^{D} \bbQ_j^{(m-1)}(\bbi_m(j), k) \\
    \notag
    &= \left( \sum_{k=1}^K \prod_{j=1}^{D} \bbQ_j^{(m-1)}(\bbi_m(j), k) \right) - \alpha^{(m)} w_0(\ccalQ^{(m)}_{1:d}, \bbi_m),
\end{align}
where $w_0$ is a polynomial function of $\ccalQ^{(m)}_{1:d} = \{ \ccalQ^{(m)}_j \}_{j=1}^d$ and is therefore bounded due to Ass. \ref{ass::bounded_iterates}. 
The last equality follows from tedious but trivial rearrangements, with details provided in the online of this paper \cite{rozada2025solving}.
Similarly, $v'_{d, m}$ refers to $\tenbQ_d^{(m)}(\bbi'_m)$. 
Next, $\nabla_{\bbQ_d} \tenbQ_d^{(m)}(\bbi_m)$ can be rewritten as
\begin{align}
    \label{eq::v_grad_tag}
    \bbu_{d, m} &= \bigodot_{j=1}^{d-1} \bbQ_j^{(m)}(\bbi_m(j), :) \bigodot_{j=d+1}^{D} \bbQ_j^{(m-1)}(\bbi_m(j), :) \\
    \notag
    &=\left( \bigodot_{j=1 \neq d}^{D} \bbQ_j^{(m-1)}(\bbi_m(j), :) \right) - \alpha^{(m)} w_1( \ccalQ^{(m)}_{1:d}, \bbi_m),
\end{align}
where $w_1$ is a bounded polynomial function, and details of the derivation are available in \cite{rozada2025solving}. Again, $\bbu'_{d, m}$ refers to the gradient $\nabla_{\bbQ_d} \tenbQ_d^{(m)}(\bbi'_m)$. With these definitions in place, we expand the stochastic gradient of the S-BCGD-PE update from \eqref{eq::stochastic_gradient} and combine it with \eqref{eq::v_tag} and \eqref{eq::v_grad_tag} to show
\begin{align}
    \notag
    &\tilde g_d(\ccalQ_d^{(m)} , \hat \sigma_m) \\
    \notag
    &= 2 \tilde \delta_{\ccalQ_d^{(m)}}(\hat \sigma_m) \nabla_{\bbQ_d} (\tenbQ_d^{(m)}(\bbi'_m) - \tenbQ_d^{(m)}(\bbi_m)) \\
    \notag
    &= 2(\hat r_m  + v'_{d,m}  - v_{d, m})  \left( \bbe_{\bbi'_m(d)} \bbu'_{d, }  - \bbe_{\bbi_m(d)} \bbu_{d, m} \right)  \\
    \notag
    &= 2 \Bigg( \hat r_m +  \sum_{k=1}^K \prod_{j=1}^D \bbQ_j^{(m-1)}(\bbi'_m(j), k) - \alpha^{(m)} w_0(\ccalQ^{(m)}_{1:d}, \bbi'_m) \nonumber \\
    \notag
    &\hspace{1.35cm}- \sum_{k=1}^K \prod_{j=1}^{D} \bbQ_j^{(m-1)}(\bbi'_m(j), k)  + \alpha^{(m)} w_0(\ccalQ^{(m)}_{1:d}, \bbi_m) \Bigg) \\
    \notag
    &\hspace{0.6cm} \left( \bbe_{\bbi'_m(d)} \!\! \left( \bigodot_{j=1 \neq d}^{D} \! \bbQ_j^{(m-1)}(\bbi'_m(j), :) \! - \! \alpha^{(m)} w_1(\ccalQ^{(m)}_{1:d}, \bbi'_m) \right) \right. \\
    \notag
    &\hspace{0.6cm} \left. - \bbe_{\bbi_m(d)} \!\! \left( \bigodot_{j=1 \neq d}^{D} \! \bbQ_j^{(m-1)}(\bbi_m(j), :) \! - \! \alpha^{(m)} w_1(\ccalQ^{(m)}_{1:d}, \bbi_m) \right) \!\! \right) \!\!.
\end{align}
Grouping terms with and without $\alpha^{(m)}$ yields
\begin{align}
    \notag
    &\tilde g_d(\ccalQ_d^{(m)}, \hat \sigma_m) \\
    \notag
    &= 2 \left( \delta_{\ccalQ^{(m-1)}}(\hat \sigma_m) - \alpha^{(m)} \left( w_0(\ccalQ^{(m)}_{1:d}, \bbi'_m) - w_0(\ccalQ^{(m)}_{1:d}, \bbi_m) \right)\right) \\
    \notag
    & \hspace{0.7cm} \Big( \nabla_{\bbQ_d} \left(\tenbQ^{(m-1)}(\bbi'_m) - \tenbQ^{(m-1)}(\bbi_m) \right)  \\
    \notag
    & \hspace{0.65cm} - \alpha^{(m)} \left( \bbe_{\bbi'_m(d)} w_1(\ccalQ^{(m)}_{1:d}, \bbi'_m) - \bbe_{\bbi_m(d)}   w_1(\ccalQ^{(m)}_{1:d}, \bbi_m) \Big)  \right) \\
    \label{eq::alpha_grad_sto}
    &= \tilde g_d(\ccalQ^{(m-1)}, \hat \sigma_m) - \alpha^{(m)} W_0 (\ccalQ^{(m)}_{1:d}, \hat \sigma_m),
\end{align}
where $W_0$ is a polynomial function which depends on the selection vectors $\bbe$, and the functions $w_0$ and $w_1$. Therefore, $W_0$ depends on the iterates $\ccalQ^{(m)}_j$ for all $j=1,\hdots, d$ and the sample $\hat \sigma_m$.
Analogously, for the gradient $g_d(\ccalQ_d^{(m)})$ we can derive
\begin{align}
    \label{eq::alpha_grad}
    g_d(\ccalQ_d^{(m)}) &= g_d(\ccalQ^{(m-1)}) - \alpha^{(m)} W_1 (\ccalQ^{(m)}_{1:d}, \hat \sigma_m),
\end{align}
where again $W_1$ is just a polynomial function that depends on $\ccalQ^{(m)}_j$ for $j=1, \hdots, d-1$ and the transition $\hat \sigma_m$. 
Therefore, equations \eqref{eq::alpha_grad} and \eqref{eq::alpha_grad_sto} imply that the gradients at iteration $m$, after updating $d-1$ blocks, can be expressed as the gradients at iteration $m-1$ plus a bounded function of order $\alpha^{(m)}$.

To prove Ass. \ref{ass::sto_bias_bound}, from \eqref{eq::alpha_grad} and \eqref{eq::alpha_grad_sto} it follows that
\begin{align}
    &\left\| \mathbb{E}_{\hat \sigma_m} \left[\tilde g_d(\ccalQ_d^{(m)}, \hat \sigma_m) - g_d(\ccalQ_d^{(m)}) \mid \sigma_{m-1} \right] \right\| \\
    &= \Big\| \mathbb{E}_{\hat \sigma_m} \Big[ \tilde g_d(\ccalQ^{(m-1)}, \hat \sigma_m) - \alpha^{(m)} W_0 (\ccalQ^{(m)}_{1:d}, \hat \sigma_m) \nonumber \\
    \notag
    &\quad - g_d(\ccalQ^{(m-1)}) + \alpha^{(m)} W_1 (\ccalQ^{(m)}_{1:d}, \hat \sigma_m) \mid \sigma_{m-1} \Big] \Big\| \\
    \notag
    &\leq \Big\| \mathbb{E}_{\hat \sigma_m} \left[\tilde g_d(\ccalQ^{(m-1)}, \hat \sigma_m) - g_d(\ccalQ^{(m-1)}) \mid \sigma_{m-1}\right] \Big\|  \nonumber \\
    \notag
    &\quad + \alpha^{(m)} \Big\| \mathbb{E}_{\hat \sigma_m} \left[W_0 (\ccalQ^{(m)}_{1:d}, \hat \sigma_m) - W_1 (\ccalQ^{(m)}_{1:d}, \hat \sigma_m) \mid \sigma_{m-1}\right] \Big\|,
\end{align}
where the last inequality follows from using the linearity of expectation and the triangular inequality.
Since $W_0$ and $W_1$ are polynomials of $\ccalQ^{(m)}_d$, bounded for all $d$ and $m$ as per Ass. \ref{ass::bounded_iterates}, there exists a finite constant $B \geq 0$ such that
\begin{align}
    &\left\| \mathbb{E}_{\hat \sigma_m} \left[\tilde g_d(\ccalQ_d^{(m)}, \hat \sigma_m) - g_d(\ccalQ_d^{(m)}) \mid \sigma_{m-1}\right] \right\| \\
    \notag
    &\leq \Big\| \mathbb{E}_{\hat \sigma_m} \left[\tilde g_d(\ccalQ^{(m-1)}, \hat \sigma_m) - g_d(\ccalQ^{(m-1)}) \mid \sigma_{m-1}\right] \Big\|  + B \alpha^{(m)} \!.
\end{align}
Recall now that S-BCGD-PE samples from an ER buffer. Therefore, $\sigma_{m}$ and $\sigma_{m-1}$ are independent. Since $\ccalQ^{(m-1)}$ does not depend on $\hat \sigma_m$ we have
\begin{align}
    \notag
    &\mathbb{E}_{\hat \sigma_m} \left[\tilde g_d(\ccalQ^{(m-1)}, \hat \sigma_m) - g_d(\ccalQ^{(m-1)}) \mid \sigma_{m-1}\right] \\
    \notag
    &= \mathbb{E}_{\hat \sigma_m} \left[\tilde g_d(\ccalQ^{(m-1)}, \hat \sigma_m)  \mid \sigma_{m-1}\right] \! - \! \mathbb{E}_{\hat \sigma_m} \! \left[g_d(\ccalQ^{(m-1)}) \! \mid \sigma_{m-1}\right] \\
    \notag
    &= g_d(\ccalQ^{(m-1)}) - g_d(\ccalQ^{(m-1)}) = 0.
\end{align}
Thus, we have
\begin{align}
    &\left\| \mathbb{E}_{\hat \sigma_m} \left[\tilde g_d(\ccalQ_d^{(m)}, \hat \sigma_m) - g_d(\ccalQ_d^{(m)}) \mid \sigma_{m-1}\right] \right\| \leq  B \alpha^{(m)}.
\end{align}
which verifies Ass. \ref{ass::sto_bias_bound}, and therefore concludes the proof.

%%%%%%%%%%%%%%%%%%%%%%%%%%%%%%%%%%%%%%%%%%

%%%%%%%%%%%%%%%%%%%%%%%%%%%%%%%%%%%%%%%%%%
\bibliographystyle{ieeetr}
\bibliography{references}

\end{document}